\documentclass[letterpaper]{article}
\usepackage{aaai21}
\usepackage{times}
\usepackage{helvet}
\usepackage{courier}
\usepackage{amsmath}
\usepackage{amssymb}
\usepackage{mathrsfs}
\usepackage{graphicx}
\usepackage{algorithm}
\usepackage{subfigure}
\usepackage{epstopdf}
\usepackage{algorithmic}
 \usepackage{natbib}
 \usepackage[switch]{lineno}
 \usepackage{color}
\newcommand{\EE}{\mathbb{E}}
\newcommand{\RR}{\mathbb{R}}
\newcommand{\PP}{\textbf{Proj}}

\usepackage{theorem}
\newtheorem{lemma}{Lemma}

\newtheorem{theorem}{Theorem}

\newtheorem{definition}{Definition}
\newtheorem{assumption}{Assumption}
\newtheorem{proposition}{Proposition}

\begin{document}
%
\title{
Stability and Generalization of the Decentralized Stochastic Gradient Descent\thanks{This work  is sponsored in part by National Key R\&D Program of China (2018YFB0204300), and the National Science Foundation of China (No. 61932001 and 61906200).}
}

\author{Tao Sun\textsuperscript{\rm 1},   Dongsheng Li\textsuperscript{\rm 1}\thanks{Corresponding author.}, and Bao Wang\textsuperscript{\rm 2}\\
\textsuperscript{\rm 1}College  of Computer, National University of Defense Technology, Changsha, Hunan, China.\\
\textsuperscript{\rm 2}Scientific Computing \& Imaging   Institute, University of Utah, USA. \\
nudtsuntao@163.com, dsli@nudt.edu.cn, wangbaonj@gmail.com
}
\maketitle

\begin{abstract}
The stability and generalization of stochastic gradient-based methods provide valuable insights into understanding the algorithmic performance of machine learning models. As the main workhorse for deep learning, stochastic gradient descent has received a considerable amount of studies. Nevertheless, the community paid little attention to its decentralized variants. In this paper, we provide a novel formulation of the decentralized stochastic gradient descent. Leveraging this formulation together with (non)convex optimization theory, we establish the first stability and generalization guarantees for the decentralized stochastic gradient descent. Our theoretical results are built on top of a few common and mild assumptions and reveal that the decentralization deteriorates the stability of SGD for the first time. We verify our theoretical findings by using a variety of decentralized settings and benchmark machine learning models.
\end{abstract}


\section{Introduction}
The great success of deep learning \citep{lecun2015deep} gives impetus to the development of stochastic gradient descent (SGD) \citep{robbins1951stochastic} and its variants \citep{nemirovski2009robust,duchi2011adaptive,rakhlin2012making,kingma2014adam,wang2020scheduled}. Although the convergence results of SGD are abundant, the effects caused by the training data is absent. To this end, the generalization error \cite{hardt2015train,lin2016generalization,bousquet2002stability,bottou2008tradeoffs} is developed as an alternative method to analyze
SGD. The generalization bound reveals the performance of stochastic algorithms
and characterizes how the training data and stochastic algorithm jointly affect the target machine learning model. To mathematically describe generalization,  \citet{hardt2015train,bousquet2002stability,elisseeff2005stability} introduce the algorithmic stability for SGD, which mainly depends on the landscape of the underlying loss function, to study the generalization bound of SGD. The stability theory of SGD has been further developed \citep{charles2018stability,kuzborskij2018data,lei2020fine}.

SGD has already been widely used in parallel and distributed settings \citep{agarwal2011distributed,dekel2012optimal,recht2011hogwild}, e.g., the decentralized SGD (D-SGD) \citep{ram2010distributed,lan2020communication,srivastava2011distributed,lian2017can}. D-SGD is implemented without a centralized parameter server, and all nodes are connected through an undirected graph. Compared to the centralized SGD, the decentralized one requires much less communication with the busiest node \citep{lian2017can}, accelerating the whole computational system.

From the theoretical viewpoint, although there exist plenty of convergence analysis of D-SGD \citep{sirb2016consensus,lan2020communication,lian2017can,lian2018asynchronous},
the stability and generalization analysis of D-SGD remains rare.


\subsection{Contributions}
In this paper, we establish the first theoretical result on the stability and generalization of the D-SGD.                                We elaborate on our contributions below.
\begin{enumerate}
\item \textit{Stability of D-SGD}:
We provide the uniform stability of D-SGD in the general convex, strongly convex, and nonconvex cases. Our theory shows that besides the learning rate, data size, and iteration number,  the stability and generalization of D-SGD are also dependent on the connected graph structure.
To the best of our knowledge, our result is the first theoretical stability guarantee for D-SGD. In the general convex setting,  we also present the stability of D-SGD in terms of the ergodic average instead of the last iteration for the excess generalization analysis.




\item \textit{Computational errors for D-SGD with convexity and projection}: We consider more general schemes of D-SGD, that is, D-SGD with projection. In the previous work \citep{ram2010distributed}, to get the convergence rate, the authors need to make additional assumptions on the graph ([Assumptions 2 and 3, \citep{ram2010distributed}]). In this paper, we remove these assumptions, and we present the computational errors of D-SGD with projections in the strongly convex setting.

\item \textit{Generalization bounds for D-SGD with convexity}: We derive (excess) generalization bounds for convex D-SGD. The excess generalization is controlled by the computational error and the generalization bound, which can be directly obtained from the stability.


\item \textit{Numerical results}: We numerically verify our theoretical results by using various benchmark machine learning models, ranging from strongly convex and convex to nonconvex settings, in different decentralized settings.

\end{enumerate}

\section{Prior Art}
In this section, we briefly review two kinds  of related works: decentralized optimization and stability and generalization analysis of SGD.

\medskip

\noindent\textbf{Decentralized  and distributed optimization}
Decentralized algorithms arise in calculating the mean of data distributed over multiple sensors \citep{boyd2005gossip,olfati2007consensus}. The decentralized (sub)gradient descent (DGD) algorithms are propose and studied by \cite{nedic2009distributed,yuan2016convergence}. Recently, DGD has been generalized to the stochastic settings. With a local Poisson clock assumption on each agent, \citet{ram2010asynchronous2} proposes an asynchronous gossip algorithm.
The decentralized algorithm with a  random communication graph is proposed in \citep{srivastava2011distributed,ram2010distributed}.
\citet{sirb2016consensus,lan2020communication,lian2017can} consider the randomness caused by the stochastic gradients and proposed the decentralized SGD (D-SGD).
The complexity analysis of D-SGD has been done in  \cite{sirb2016consensus}.
In \citep{lan2020communication}, the authors propose another kind of D-SGD that leverages dual information, and provide the related computational complexity. In the paper \citep{lian2017can}, the authors show the advantage of D-SGD compared to the centralized SGD.
In a recent paper \citep{lian2018asynchronous}, the authors developed asynchronous D-SGD with theoretical convergence guarantees. The biased decentralized SGD is proposed and studied by \citep{sun2019decentralized}.   In  \cite{richards2020graph}, the authors studied the stability for a non-fully decentralized training method, in which each node needs to communicate extra gradient information. Paper \cite{richards2020graph} is closed to ours,  but we consider the DSGD, which is   different from the  algorithm investigated by \cite{richards2020graph} and more general. Further more, we studied the nonconvex settings.

\medskip

\noindent \textbf{Stability  and Generalization   of SGD} In \citep{shalev2010learnability}, on-average stability is proposed and further studied by \citet{kuzborskij2018data}.  The uniform
stability of empirical risk minimization (ERM) under strongly convex objectives is considered by \citet{bousquet2002stability}.  Extended results are proved with the pointwise-hypothesis assumption, which shows that a class of learning algorithms is convergent with global optimum \citep{charles2018stability}. In order to prove uniform stability of SGD, \citet{hardt2015train} reformulate SGD as a contractive iteration. In \citep{lei2020fine}, a new stability notion is proposed to remove the bounded gradient assumptions. In \citep{bottou2008tradeoffs}, the authors  establish a framework for the generalization performance of SGD. \citet{hardt2015train} connects the uniform stability with generalization error. The generalization errors with strong convexity are established in \cite{hardt2015train,lin2016generalization}. The stability and generalization are also studied for the Langevin dynamics \citep{li2019generalization,mou2018generalization}.

\medskip

 \section{Setup}
This part contains preliminaries and mathematical descriptions of our problem.
Analyzing the stability of D-SGD is more complicated than that of SGD due to the challenge arises from the mixing matrix in D-SGD. We cannot directly adapt the analysis for SGD to D-SGD. To this end, we reformulate D-SGD as an operator iteration with an error term, which is followed by bounding the error in each iteration.


  \subsection{Stability and Generalization }
 The   population risk minimization  is an important model in machine learning and statistics, whose mathematical formulation reads as
 \begin{align*}
 \min_{{\bf x}\in\RR^d}R({\bf x}):=\EE_{\xi\sim \mathcal{D}} f({\bf x}; \xi),
 \end{align*}
 where $f({\bf x}; \xi)$ denotes the loss of the model associated with data $\xi$ and $\mathcal{D}$ is the data distribution. Due to the fact that $\mathcal{D}$ is usually unknown or very complicated, we consider the following surrogate ERM
  \begin{align*}
  \min_{{\bf x}\in\RR^d} R_S({\bf x}):= \frac{1}{N}\sum_{i=1}^N f({\bf x}; \xi_i),
 \end{align*}
 where  $S:=\{\xi_1,\xi_2,\ldots,\xi_N\}$ and $~\xi_i\sim \mathcal{D}$ is a given data.

 \medskip

For a specific stochastic algorithm $\mathcal{A}$ act on $S$ with output $\mathcal{A}(S)$, the \textit{generalization error} of $\mathcal{A}$ is defined as
  $\epsilon_{\textrm{gen}}:=\EE_{S,\mathcal{A}}[R(\mathcal{A}(S))-R_{S}(\mathcal{A}(S))].$
  Here, the expectation is taken over 
  the algorithm and the data.
  The generalization  bound reflects the joint effects caused by the data $S$ and the algorithm $\mathcal{A}$.
 We are also interested in the \textit{excess generalization error}, which is defined as $\epsilon_{\textrm{ex-gen}}:=\EE_{S,\mathcal{A}}[R(\mathcal{A}(S))-R({\bf x}^*)]$, where ${\bf x}^*$ is the minimizer of $R$.
 Let $\overline{{\bf x}}$ be the minimizer of $R_{S}$.
  Due to the unbiased expectation of the data $S$, we have  $\EE_{S}[R_{S}({\bf x}^*)]=\EE [R({\bf x}^*)]$. Thus,  \citet{bottou2008tradeoffs} point out $\epsilon_{\textrm{ex-gen}}$ can be decomposed as follows
  \begin{equation*}
  \begin{aligned}
  &\EE_{S,\mathcal{A}}[R(\mathcal{A}(S))-R({\bf x}^*)]=\underbrace{\EE_{S,\mathcal{A}}[R(\mathcal{A}(S))-R_{S}(\mathcal{A}(S))]}_{\textrm{generalization  error}}\\
  &+\underbrace{\EE_{S,\mathcal{A}}[R_{S}(\mathcal{A}(S))-R_{S}(\overline{{\bf x}})]}_{\textrm{optimization error}}+\underbrace{\EE_{S,\mathcal{A}}[R_{S}(\overline{{\bf x}})-R_{S}({\bf x}^*)]}_{\textrm{test error}}.
  \end{aligned}
  \end{equation*}
 Notice that $R_{S}(\overline{{\bf x}})\leq R_{S}({\bf x}^*)$, therefore
 $$\epsilon_{\textrm{ex-gen}}\leq \epsilon_{\textrm{gen}}+\EE_{S,\mathcal{A}}[R_{S}(\mathcal{A}(S))-R_{S}(\overline{{\bf x}})].$$

\medskip

The \textit{uniform stability} is used to bound the generalization error of a given algorithm $\mathcal{A}$ \citep{hardt2015train,elisseeff2005stability}.
\begin{definition}
We say  that the randomized algorithm $\mathcal{A}$ is $\epsilon$-\emph{uniformly stable} if for any two data sets $S,S'$ with $n$ samples that differ in one example, we have
\begin{equation*}
\sup_{\xi} \EE_{\mathcal{A}} \left[ f(\mathcal{A}(S); \xi) - f(\mathcal{A}(S'); \xi) \right] \le \epsilon.
\end{equation*}
 \end{definition}
It has been proved that  the uniform  stability directly implies the generalization bound.
 \begin{lemma}[\cite{hardt2015train}]
Let $\mathcal{A}$ be $\epsilon$-\emph{uniformly stable}, it follows
$
|\EE_{S,\mathcal{A}}[R(\mathcal{A}(S))-R_{S}(\mathcal{A}(S))]| \leq\epsilon.
$
 \end{lemma}
Thus, to get the generalization  bound of a random algorithm, we just need to compute the uniform stability bound $\epsilon$.

  \subsection{Problem Formulation }
{\bf Notation}: We use the following notations throughout the paper. We denote the $\ell_2$ norm of ${\bf x}\in \RR^d$ as $\|{\bf x}\|$.
For a matrix ${\bf A}$, ${\bf A}^{\top}$ denotes its transpose, we denote the spectral norm of ${\bf A}$ as $\|{\bf A}\|_{\textrm{op}}$. Given another matrix ${\bf B}$, ${\bf A} \succ {\bf B}$ means that ${\bf A}-{\bf B}$ is positive define; and ${\bf A} \succeq {\bf B}$ means ${\bf A}-{\bf B}$ is positive semidefinite. The identity matrix is defined as $\mathbb{I}$. We use $\EE[\cdot]$ to denote the expectation of $\cdot$ with respect to the underlying probability space. For two positive constants $a$ and $b$, we denote $a=\mathcal{O}(b)$ if there exists $C>0$ such that $a\leq C b$, and
$\widetilde{\mathcal{O}}(b)$ hides a logarithmic factor of $b$.

\medskip

\noindent 
{ Let   $\mathcal{D}_i=\{\xi_{l(i)}\}_{1\leq l\leq n}$ ($1\leq i\leq m$) denote  the data stored  in the $i$th client, which follow the same distribution of $\mathcal{D}$ \footnote{For simplicity, we assume all clients have the same amount of samples.}.  }
In this paper, we consider solving the objective function ~\eqref{dec} by the DGD, where
\begin{align}\label{dec}
  f({\bf x}):=\frac{1}{mn}\sum_{i=1}^{m} \sum_{ l=1}^{n} f({\bf x}; \xi_{l(i)}).
\end{align}
Note that \eqref{dec} is a decentralized approximation to the following population risk function
\begin{align}\label{po}
  F({\bf x}):=\EE_{\xi\sim\mathcal{D}}f({\bf x}; \xi).
\end{align}
To distinguish from the objective functions in the last subsection, we use $f$ rather than $R_{S}$ here. The decentralized optimization is usually associated with a mixing matrix, which is designed by the users according to a given graph structure. In particular, we consider the connected graph $\mathcal{G} = (\mathcal{V}, \mathcal{E})$ with vertex set $\mathcal{V}=\{1,...,M\}$ and edge set $\mathcal{E}\subseteq \mathcal{V}\times \mathcal{V}$ with edge $(i, l)\in\mathcal{E}$ represents the communication link between nodes  $i$
and $l$. Before proceeding, let us recall the  definition of the mixing matrix.
\begin{definition}[Mixing matrix]
\label{def:MixMat} For any given graph $\mathcal{G}=(\mathcal{V},\mathcal{E})$,
the mixing matrix ${\bf W} = [w_{ij}] \in \mathbb{R}^{M\times M}$ is defined on the edge set $\mathcal{V}$ that satisfies:
(1) If $i\neq j$ and $(i,j) \notin {\cal E}$, then $w_{ij} =0$; otherwise, $w_{ij} >0$;
(2) ${\bf W} = {\bf W}^{\top}$;
(3) $\mathrm{null} \{\mathbb{I}-{\bf W}\} = \mathrm{span}\{\bf 1\}$;
(4) $\mathbb{I} \succeq {\bf W} \succ -\mathbb{I}.$
\end{definition}
Note that
${\bf W}$ is a doubly  stochastic matrix \citep{marshall1979inequalities}, and
the mixing matrix is non-unique for a given graph. Several common examples for ${\bf W}$ include the Laplacian matrix and  the maximum-degree matrix \citep{boyd2004fastest}. A crucial constant that characterizes the mixing matrix is
$$\lambda:=\max\{|\lambda_2|,|\lambda_m({\bf W})|\},$$
where $\lambda_i$ denotes the $i$th largest eigenvalue of ${\bf W}\in\RR^{m\times m}$.
The definition of the mixing matrix implies that $0\leq\lambda<1$.
\begin{lemma}[Corollary 1.14., \cite{montenegro2006mathematical}]\label{mi}
Let
$
    {\bf P}\in \mathbb{R}^{m\times m}
$ be the matrix whose elements are all $1/m$.
Given any $k\in \mathbb{Z}^+$, the mixing matrix ${\bf W}\in\RR^{m\times m}$ satisfies
$$\|{\bf W}^k-{\bf P}\|_{\emph{op}}\leq \lambda^k.$$
\end{lemma}
Note the fact that the stationary distribution of an irreducible aperiodic finite Markov chain is uniform if and only if its transition matrix is doubly stochastic.
Thus, ${\bf W}$ corresponds to some Markov chain's transition matrix, and the parameter $0\leq \lambda<1$ characterizes the speed of convergence to the stationary state.

\medskip

We consider a general decentralized stochastic gradient descent with projection, which carries out in the following manner: in the $t$-th iteration,
  1) client $i$ applies an approximate copy ${\bf x}^{t}(i)\in \RR^d$ to calculate a unbiased gradient estimate $\nabla f({\bf x}^{t}(i);\xi_{j_t(i)})$, where $j_t(i)\in \mathbb{Z}^+$ is the local random index;
 2) client $i$ replaces its local parameters with the weighted average of its neighbors, i.e.,
 \begin{align}\label{connected}
 \tilde{{\bf x}}^{t}(i) = \sum_{l\in \mathcal{N}(i)} w_{i,l} {\bf x}^{t}(l);
 \end{align}
   3)   client $i$ updates its parameters as
   \begin{equation}\label{connected2}
        \begin{aligned}
{\bf x}^{t+1}(i)&=\textbf{Proj}_{V}\Big( \tilde{{\bf x}}^{t}(i)-\alpha_t \nabla f({\bf x}^t(i);\xi_{j_t(i)}) \Big)
 \end{aligned}
   \end{equation}
 with learning rate $\alpha_t>0$, and ${\bf Proj}_V(\cdot)$ stands for projecting the quantity $\cdot$ into the space $V$. We stress that, in practice, we do not need to compute the average ${\bf x}^t=\frac{1}{m}\sum_{i=1}^m {\bf x}^t(i)$ in each iteration, and we take the average only in the last iteration.
\begin{algorithm}
\caption{Decentralized Stochastic Gradient Descent (D-SGD)}
\begin{algorithmic}\label{alg}
\REQUIRE   $(\alpha_t>0)_{t\geq 0} $,  initialization   ${\bf x}^0$\\
\textbf{for}~node $i=1,2,\ldots,m$ \\
~~\textbf{for}~$t=1,2,\ldots$ \\
~~~ updates local parameter  as \eqref{connected} and \eqref{connected2}\\
~~~ ${\bf x}^t=\frac{1}{m}\sum_{i=1}^m {\bf x}^t(i)$\\
~~\textbf{end for}\\
\textbf{end for}\\
\end{algorithmic}
\end{algorithm}

\medskip

In the following, we draw necessary assumptions, which are all common and widely used in the nonconvex analysis community.
\begin{assumption}\label{ass1}
The loss function $f({\bf x};\xi)$ is nonnegative and differentiable with respect to ${\bf x}$, and $\nabla f({\bf x};\xi)$ is bounded by the constant $B$ over $V$, i.e., $\max_{{\bf x}\in V,\xi\sim \mathcal{D}}\|\nabla f({\bf x};\xi)\|\leq B$.
\end{assumption}
Assumption \ref{ass1} implies that
$|  f({\bf x};\xi) -  f({\bf y};\xi)| \leq B \|{\bf x} - {\bf y}\|,$
for all ${\bf x}, {\bf y} \in V$ and any  $\xi\sim \mathcal{D}$.

\begin{assumption}\label{ass2}
The gradient of $f({\bf x};\xi)$ with respect to ${\bf x}$ is $L$-Lipschitz, i.e.,
$\| \nabla f({\bf x};\xi) -  \nabla  f({\bf y};\xi)\| \leq L \|{\bf x} - {\bf y}\|,$
for all ${\bf x}, {\bf y} \in V$ and any $\xi\sim \mathcal{D}$.

\end{assumption}

\begin{assumption}\label{ass3}
The set $V$ forms a closed ball in $\RR^d$.
\end{assumption}

Compared with the scheme presented in \citep{lian2017can}, our algorithm accommodates a projection after each update in each client. When $\nabla f({\bf x};\xi)$ is non-strongly convex, $V$ can be set as the full space and Algorithm 1 reduces to the scheme given in \citep{lian2017can}, whose convergence has been well studied. Such a projection is more general and is necessary for the strongly convex analysis; we explain this necessary claim as follows: if $f$ is $\nu$-strongly convex, then $\|\nabla f({\bf x})\|^2\geq \nu\|{\bf x}-{\bf x}^*\|^2$ with ${\bf x}^*$ being the minimizer of $f({\bf x})$ \citep{karimi2016linear}. Thus, when ${\bf x}$ is far from ${\bf x}^*$, the gradient is unbounded, which breaks Assumption \ref{ass1}. However, with the projection procedure, D-SGD (Algorithm 1) actually  minimizes function \eqref{dec} over the set $V$. The strong convexity gives us $f({\bf x})-f({\bf x}^*)\geq \frac{\nu}{2} \|{\bf x}-{\bf x}^*\|^2$, which indicates ${\bf x}^*\in \textbf{B}(\textbf{0},\sqrt{{2(f({\bf x}^0)-\min f)}/{\nu}})$. Thus, when the radius of $V$ is large enough, the projection does not change the output of D-SGD.

\section{Stability of D-SGD}
In this section, we prove the stability theory for D-SGD in strongly convex, convex, and nonconvex settings.

\subsection{General Convexity }
This part contains the stability result of D-SGD when $f(\cdot;\xi)$ is generally convex.
\begin{theorem}\label{th1}
Let $f(\cdot;\xi)$ be convex and Assumptions \ref{ass1}, \ref{ass2}, \ref{ass3} hold. If the step size $\alpha_t\leq {2}/{L}$, then D-SGD satisfies the uniform stability with
$$\epsilon_{\emph{stab}}\leq \frac{2B^2\sum_{t=1}^{T-1}\alpha_t}{mn}+4B^2\sum_{t=1}^{T-1}\Big[(1+\alpha_t B)\sum_{j=0}^{t-1}\alpha_j\lambda^{t-1-j}\Big].$$
\end{theorem}
Compared to the results of minimizing  \eqref{dec} by using centralized SGD with step sizes $(\alpha_t)_{t\geq 1}$ [Theorem 3.8, \citep{hardt2015train}], which yields the uniformly stable bound as ${2B^2\sum_{t=1}^{T-1}\alpha_t}/{(mn)}$.  Theorem \ref{th1} shows that D-SGD suffers from an additional term $4B^2\sum_{t=1}^{T-1}( 1+\alpha_t B)\sum_{j=0}^{t-1}\alpha_j\lambda^{t-1-j}$, which does not vanish when $\lambda>0$.


If we set $\alpha_t={1}/{(t+1)}$, with Lemma \ref{th-1}, it is easy to check that $\epsilon_{\textrm{stab}}=\mathcal{O}( \frac{\ln T}{mn}+C_{\lambda}\ln T)$; However, if we use a constant learning rate, (i.e., $\alpha_t\equiv \alpha$), when $0<\lambda<1$, we have $4B^2\sum_{t=1}^{T-1}( 1+\alpha_t B)\sum_{j=0}^{t-1}\alpha_j\lambda^{t-1-j}=\mathcal{O}(\frac{\alpha T}{1-\lambda})$ and $\epsilon_{\textrm{stab}}=\mathcal{O}(\frac{\alpha T}{1-\lambda}+\frac{\alpha T}{mn})$. The result indicates that although decentralization reduces the busiest node's communication, it hurts the stability.

\medskip

Theorem \ref{th1} provides the uniform stability for the last-iterate of D-SGD. However, the computational error of D-SGD in general convexity case uses the following average
\begin{equation}\label{average}
   \textrm{ave}({\bf x}^T):=\frac{\sum_{t=1}^{T-1}\alpha_t{\bf x}^t}{\sum_{t=1}^{T-1}\alpha_t}.
\end{equation}
Such a mismatch leads to the difficulty in characterizing the excess generalization bound. It is thus necessary describe to the uniform stability in terms of  $\textrm{ave}({\bf x}^T)$. To this end, we consider that D-SGD outputs $ \textrm{ave}({\bf x}^t)$ instead of ${\bf x}^t$ in the $t$-th iteration. The uniform stability, in this case, is defined as $\epsilon_{\textrm{ave-stab}}$, and we have the following result.

\begin{proposition}\label{pro1}
Let $f(\cdot;\xi)$ be convex and  Assumptions \ref{ass1}, \ref{ass2}, \ref{ass3} hold. If the step size $\alpha_t\equiv\alpha\leq {2}/{L}$, the uniform stability $\epsilon_{\emph{ave-stab}}$, in terms of $\textrm{ave}({\bf x}^t)$, satisfies
$$\epsilon_{\emph{ave-stab}} \leq \frac{2B^2\alpha(t-1)}{mn}+\frac{4\alpha B^2( 1+\alpha B)(t-1)}{1-\lambda}1_{\lambda\neq 1}.$$
Furthermore, if the step size is chosen as $\alpha_t={1}/{(t+1)}$, we have
$$\epsilon_{\emph{ave-stab}}\leq \frac{B^2\ln T}{mn}+\frac{4B^2( 1+B)}{\ln (T+1)}1_{\lambda\neq 1}.$$
\end{proposition}
Unlike the uniform stability for ${\bf x}^T$, the average turns out to be a
very complicated one. We thus just present two classical kinds of step size.

\begin{figure*}[!htb]
    \centering
  \subfigure[Random]{ \includegraphics[width=0.14\textwidth]{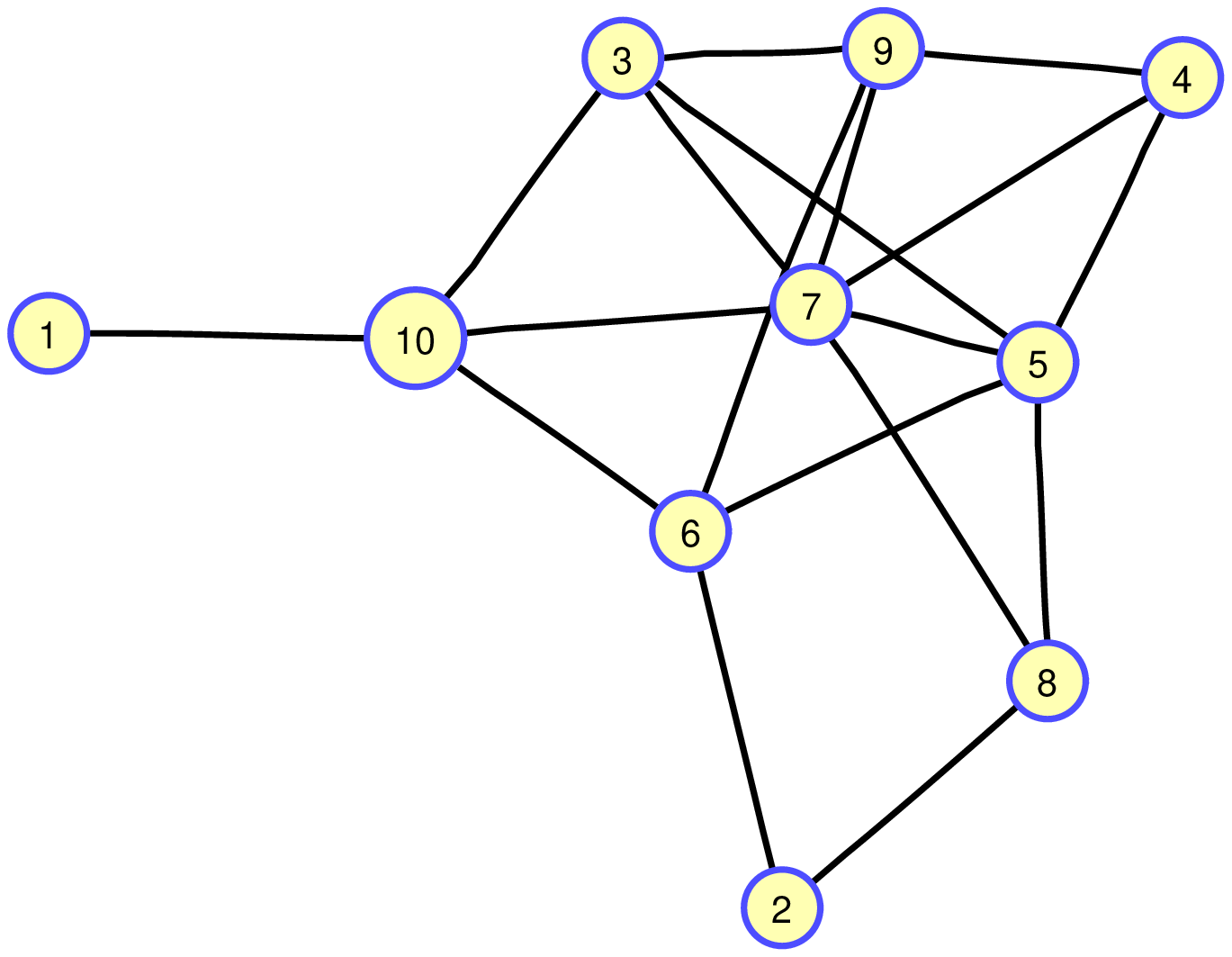}}
  \subfigure[Star]{\includegraphics[width=0.14\textwidth]{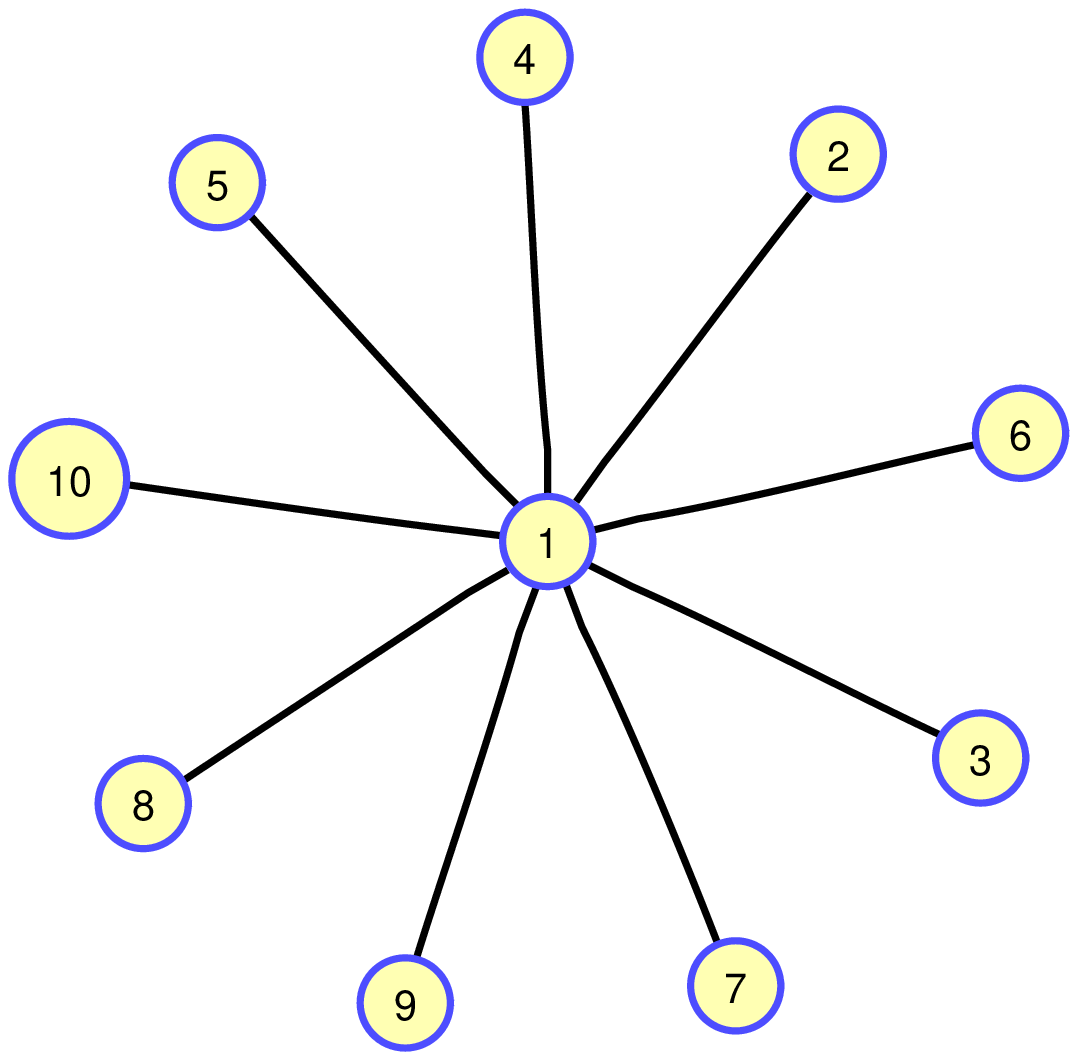}}
  \subfigure[Cycle]{ \includegraphics[width=0.14\textwidth]{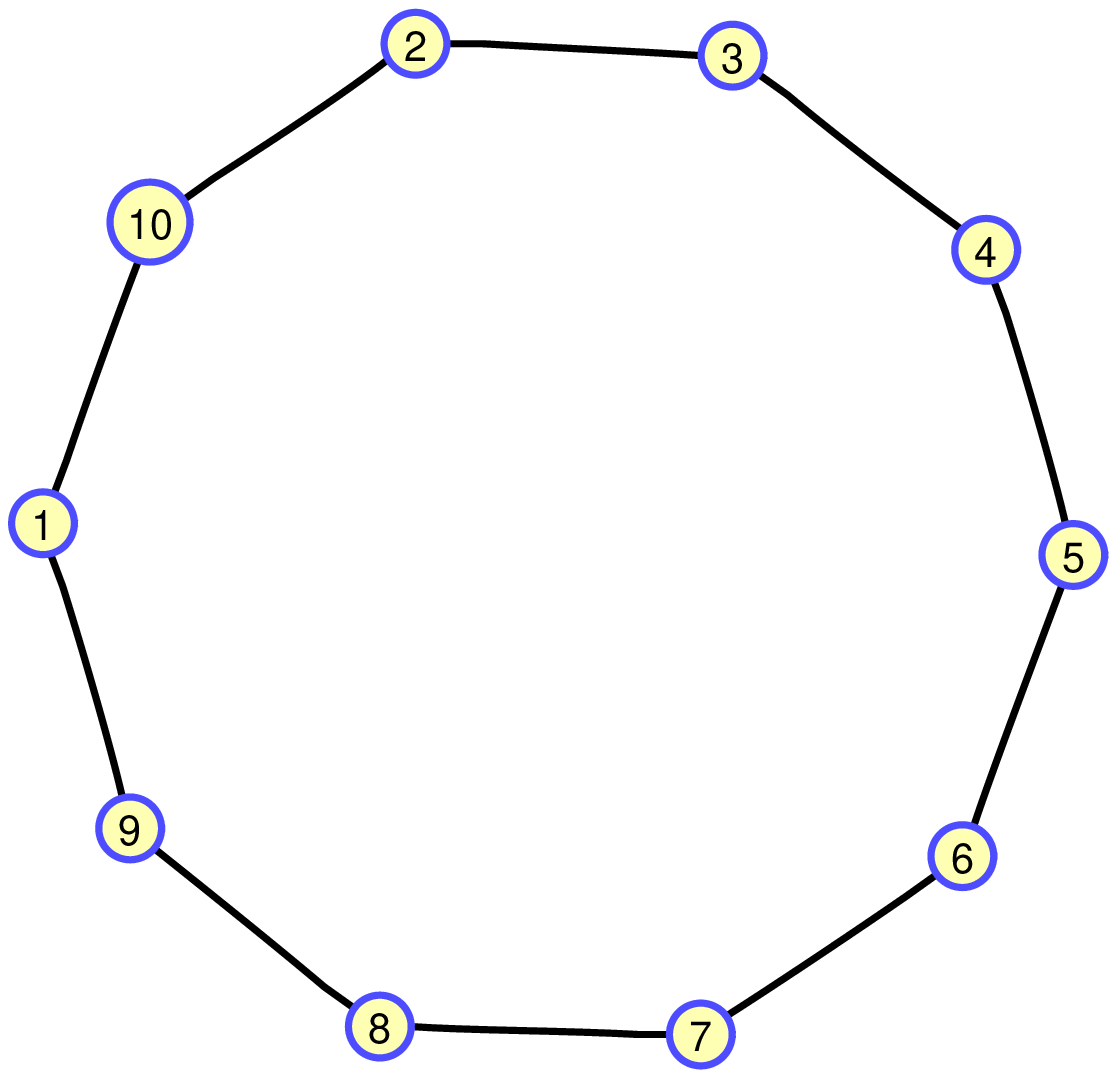}}
  \subfigure[k-NNG]{ \includegraphics[width=0.14\textwidth]{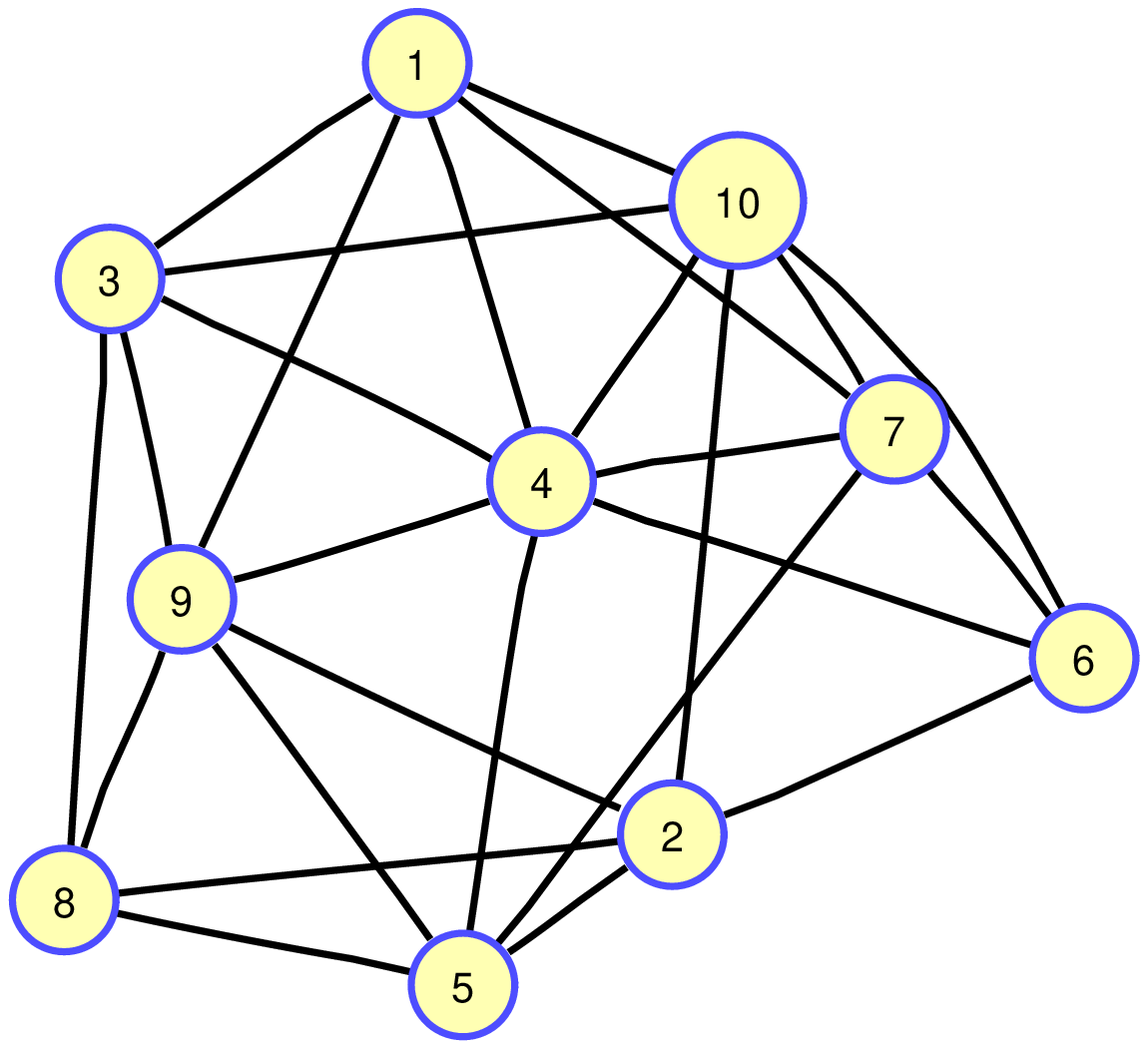}}
    \subfigure[Bipartite]{ \includegraphics[width=0.14\textwidth]{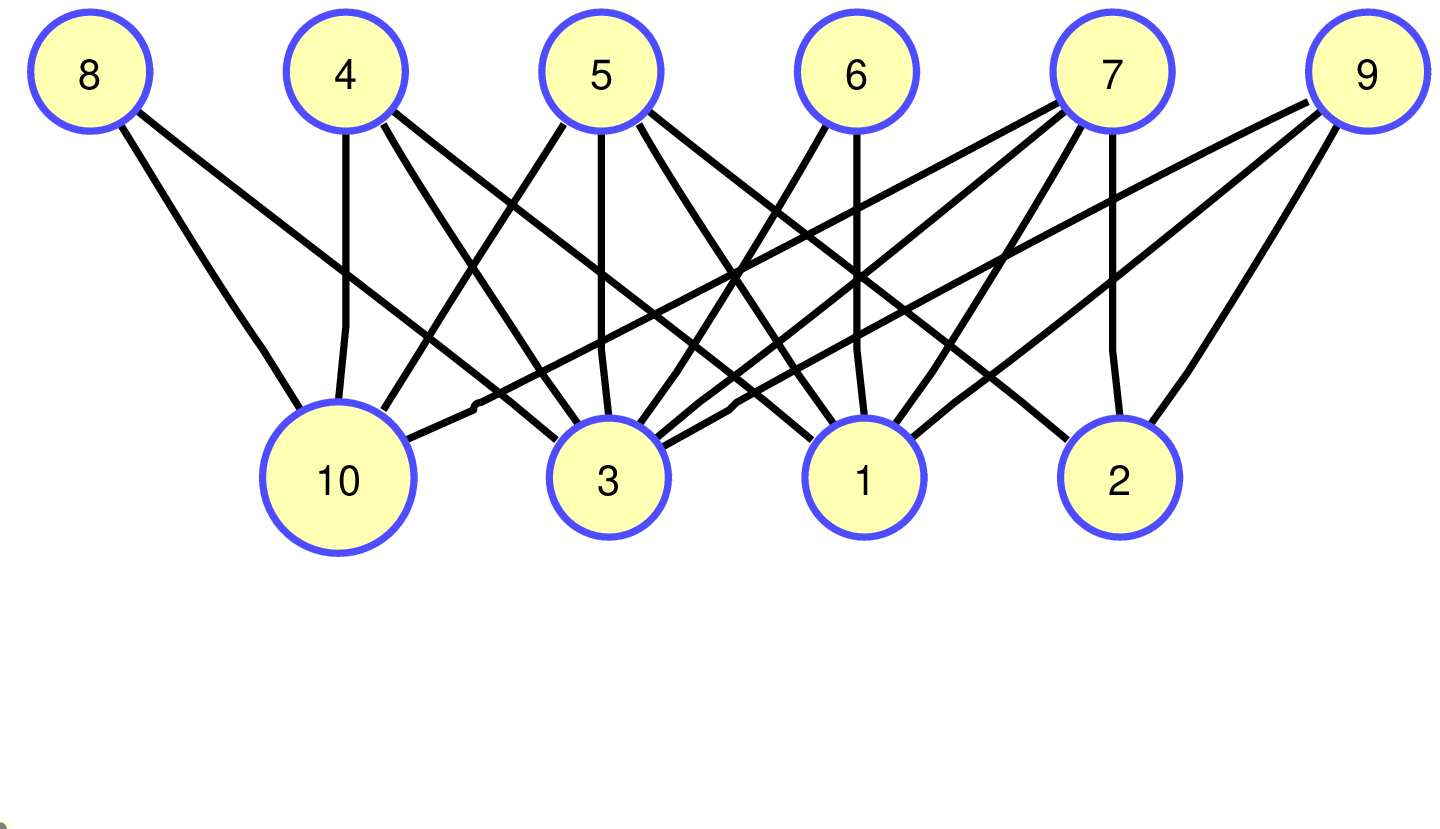}}
        \subfigure[Complete]{ \includegraphics[width=0.14\textwidth]{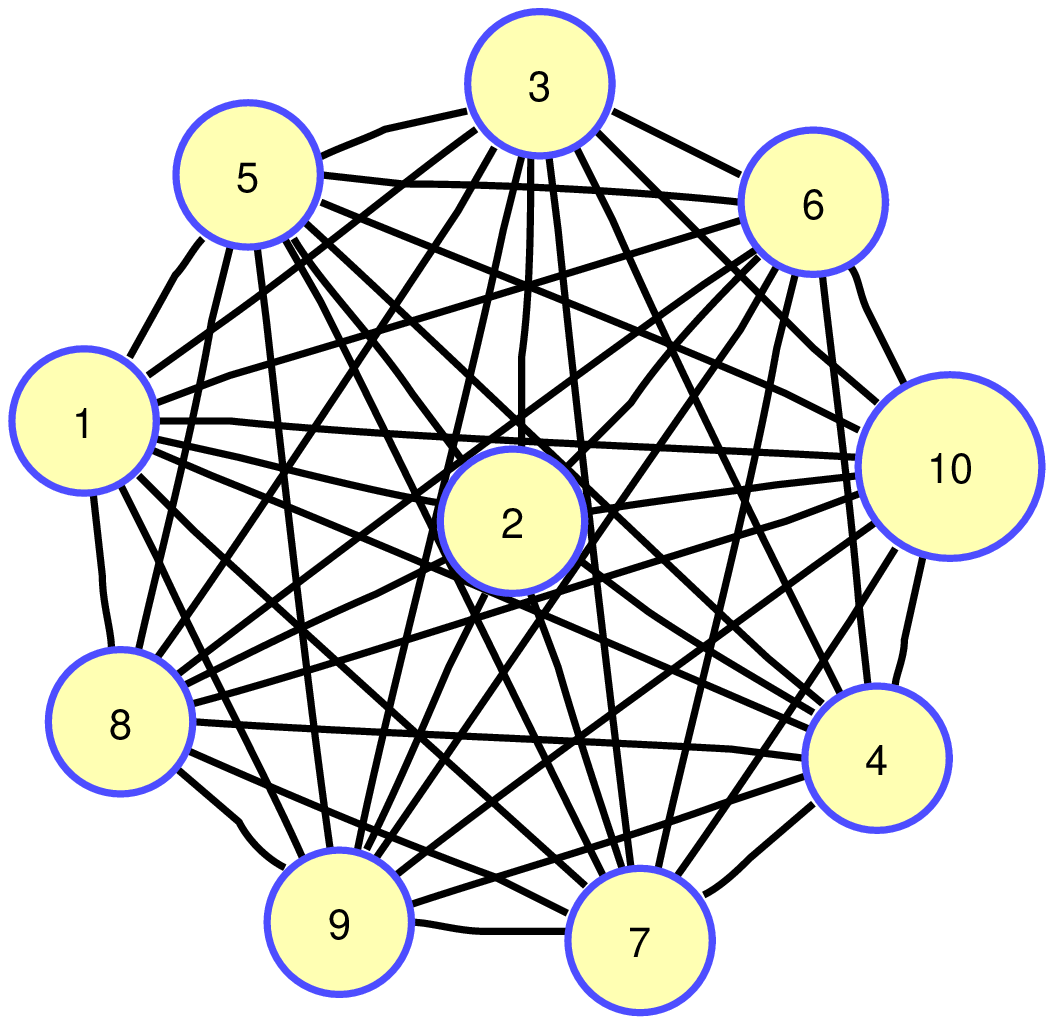}}
    \caption{ Structures of the connected graph used in our numerical tests.
    }
    \label{graphs}
\end{figure*}

 \subsection{Strong Convexity}
In the convex setting, for a fixed iteration number $T$, as the data size $mn$ increases and $\lambda$ decreases, $\epsilon_{\textrm{stab}}$ gets smaller for both diminishing and constant learning rates. However, similar to SGD, D-SGD also fails to have $\epsilon_{\textrm{stab}}$ under control when $T$ increases. This drawback does not exist in
the strongly convex setting.

Strongly convex loss functions appear in the $\ell_2$ regularized machine learning models. As mentioned in Section 2, to guarantee the bounded gradient, the set $V$ should be restricted to a closed ball. We formulate the uniform stability results in this case in Theorem~\ref{th2}.
 \begin{theorem}\label{th2}
Let $f(\cdot;\xi)$ be $\nu$-strongly convex and  Assumptions \ref{ass1}, \ref{ass2}, \ref{ass3} hold. If the step size $\alpha_t\equiv\alpha\leq {1}/{L}$, then D-SGD satisfies the uniform stability with
$$\epsilon_{\emph{stab}}\leq \frac{2B^2}{mn\nu}+\frac{4( 1+\alpha B)B^2}{\nu} \frac{1_{\lambda\neq 0}}{1-\lambda}.$$
Furthermore, if the step size $\alpha_t=\frac{1}{\nu (t+1)}$, it holds that
$$\epsilon_{\emph{stab}}\leq\frac{2B^2}{mn\nu }+ 4( 1+ \frac{ B}{\nu})\frac{ B^2}{\nu}  \frac{1_{\lambda\neq 0}}{1-\lambda}.$$
\end{theorem}
The uniformly stability bound for SGD with strong convexity is ${2B^2}/{(mn\nu)}$ \citep{hardt2015train}, which is smaller than the one of D-SGD. From Theorem \ref{th2}, we see that the uniform stability bound of D-SGD is independent  on the iterative number $T$. Moreover, D-SGD enjoys a smaller uniformly stable bound when the data size $mn$ is larger and $\lambda$ is smaller. 
 \subsection{Nonconvexity}
We now present the stability result for nonconvex loss functions.

  \begin{theorem}\label{th3}
Suppose Assumptions \ref{ass1}, \ref{ass2}, \ref{ass3} hold and $\sup_{{\bf x}\in V,\xi}f({\bf x};\xi)\leq 1$. For any $T$, if the step size $\alpha_t\leq {c}/{(t+1)}$ and $c$ is small enough, then D-SGD satisfies the uniform stability with
\begin{align*}
\epsilon_{\emph{stab}}&\leq \frac{c^{\frac{1}{1+cL}}T^{\frac{cL}{1+cL}}}{mn}\\
&+c^{\frac{1}{1+cL}}\Big[\frac{2B^2cL}{mn}+4( 1+c B)B^2LC_{\lambda}\Big]T^{\frac{cL}{1+cL}}.
\end{align*}
\end{theorem}
Without the convexity assumption, the uniform stable bound of D-SGD deteriorated. Theorem \ref{th3} shows that   $\epsilon_{\textrm{stab}}=\mathcal{O}((1+C_{\lambda}){T^{\frac{cL}{1+cL}}}/{(mn)})$, which is much larger than the bounds in the convex case $\left(\mathcal{O}({\ln T}/{(mn)}+C_{\lambda}\ln T)\right)$.

 \begin{figure}[!htb]
    \centering
  \subfigure[Linear regression ]{ \includegraphics[width=0.18\textwidth]{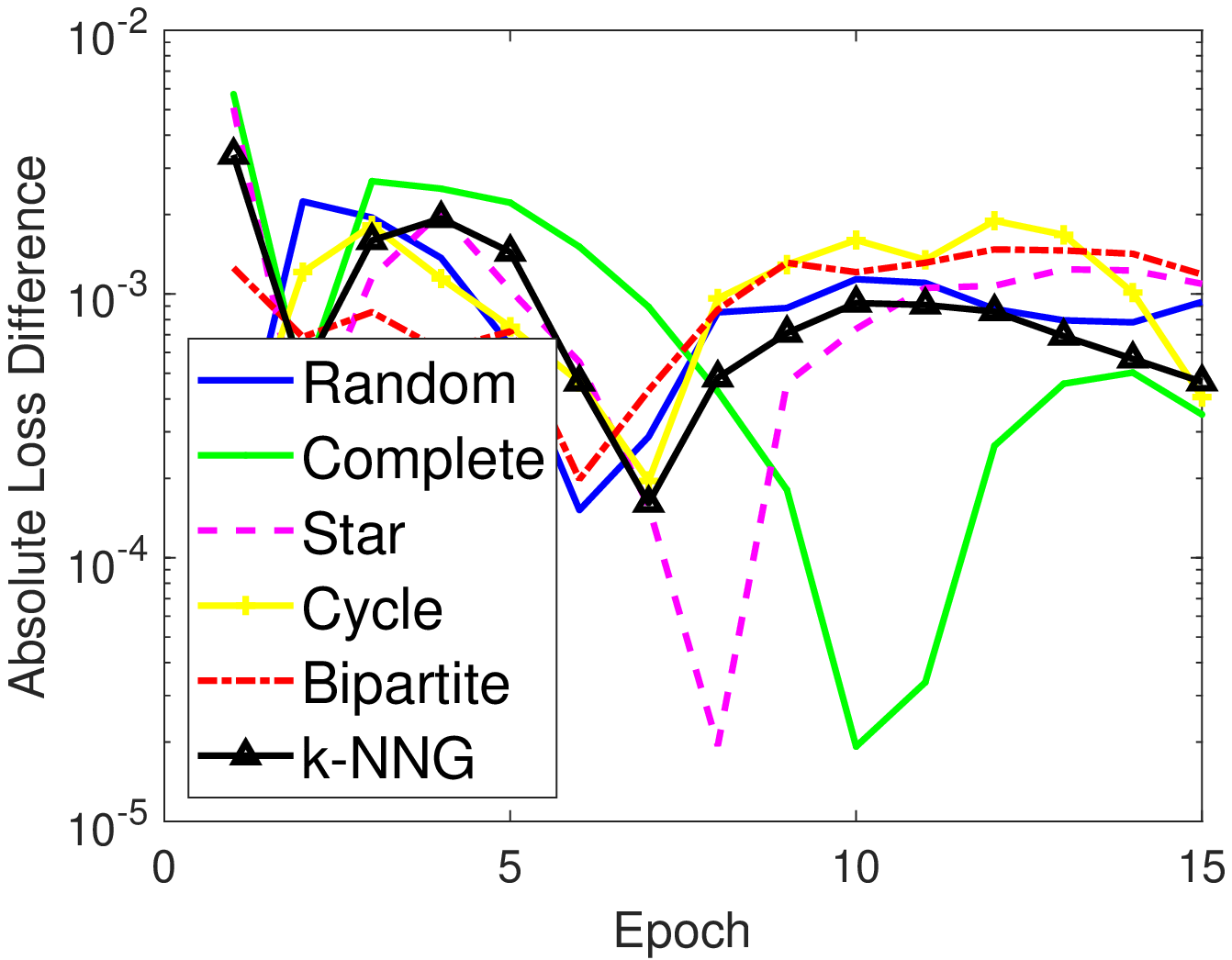}}
    \subfigure[Regularized logistic regression ]{ \includegraphics[width=0.18\textwidth]{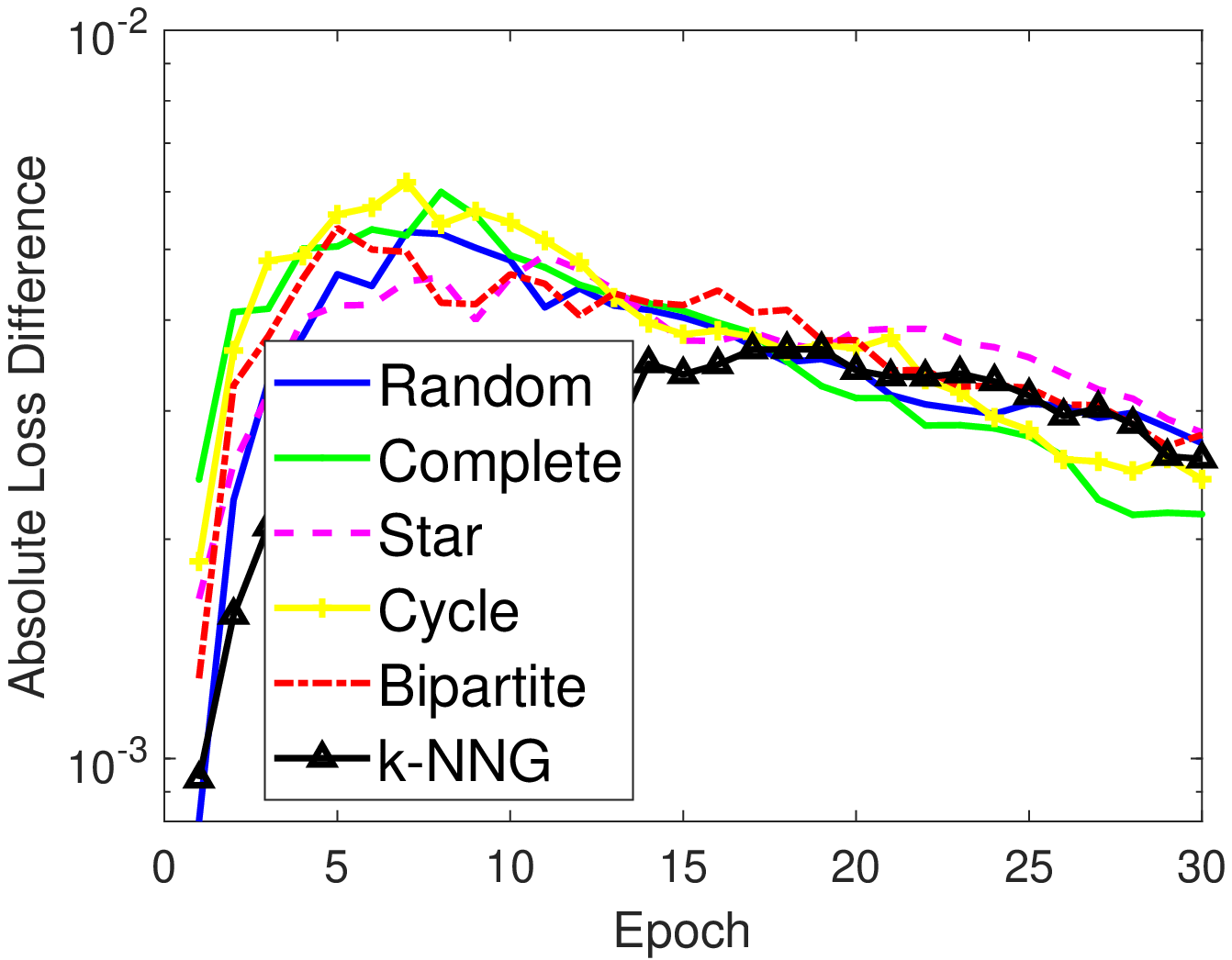}}
     \subfigure[ResNet-20 classification on Cifar-10]{ \includegraphics[width=0.18\textwidth]{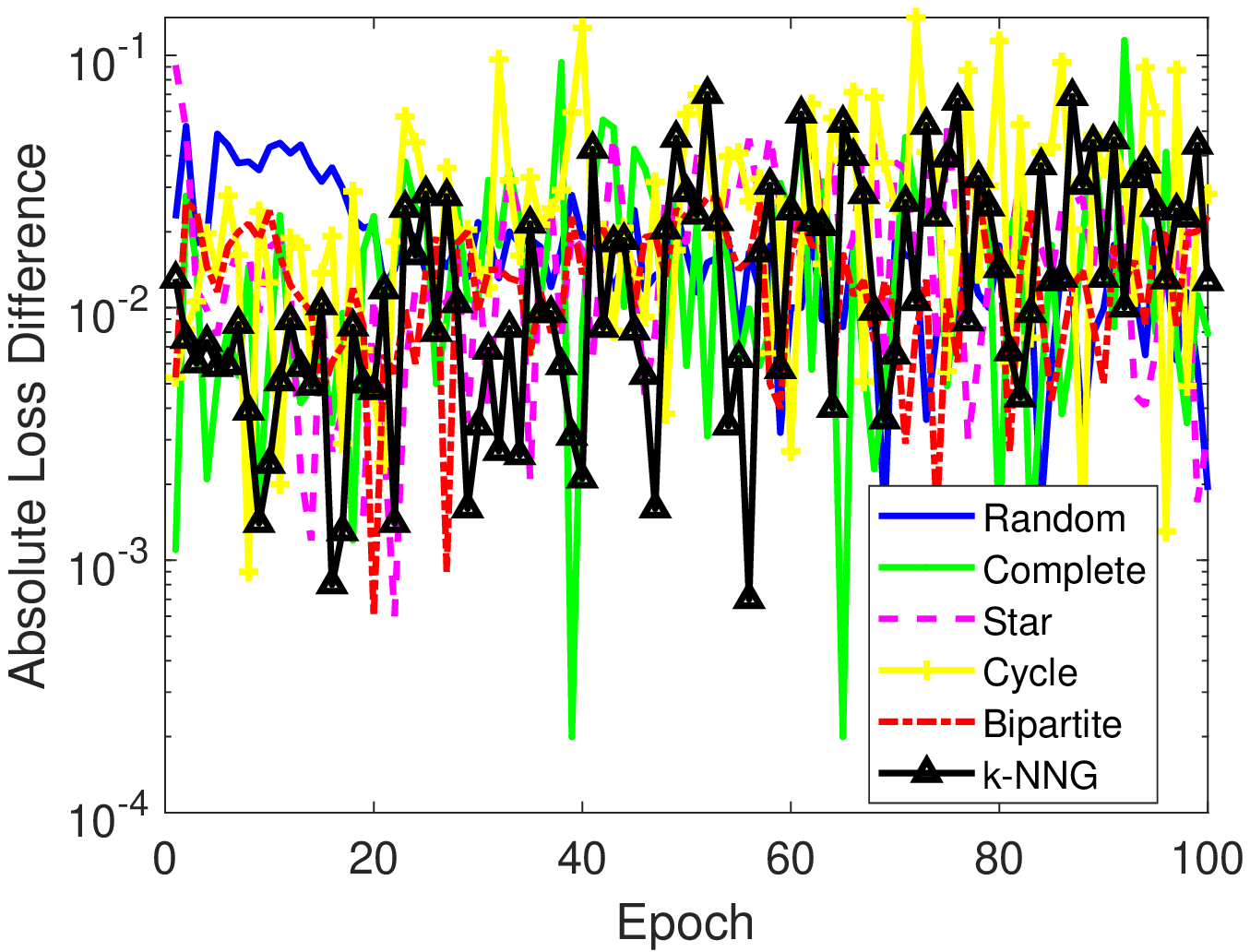}}
    \caption{\large Comparison  of the absolute loss difference under different graphs. (a), (b) and (c) correspond to the general convex, strongly convex, and nonconvex cases, respectively. In the strongly convex case, the curves become stable after enough iterations for all graphs. In the general convex case, the absolute loss difference oscillates and inferior to the strongly convex case.
    D-SGD performs worst in the nonconvex tests in terms of stability.}
    \label{comparison}
\end{figure}

\section{Excess Generalization  for Convex Problems}
In the nonconvex case, the optimization error of the function value is unclear. Thus, the excess generalization error is absent.
We are also interested in the excess generalization associated with the computational optimization error. The existing computational errors of Algorithm \ref{alg} require extra assumptions on the graph for projections. However, these assumptions may fail to hold in many applications. Thus, we first present the optimization error of D-SGD when $f({\bf x};\xi)$ is convex without extra assumptions.

 \subsection{Optimization Error of Convex D-SGD }
This part consists of optimization errors of D-SGD for convex and strongly convex settings. Assume ${\bf x}^*$ is the minimizer of $f({\bf x})$ over the set $V$, i.e., $f({\bf x}^*)=\min_{{\bf x}\in V}f({\bf x})$.

\begin{lemma}\label{th-1}
Let $f(\cdot;\xi)$ be    convex   and  Assumptions \ref{ass1}, \ref{ass2} hold, and let $({\bf x}^t)_{1\leq t\leq T}$ be the sequence generated by D-SGD. Then
\begin{align*}
\small
& \EE(f(\emph{ave}({\bf x}^T))-f({\bf x}^*)) \leq \frac{\|{\bf x}^1-{\bf x}^*\|^2}{\sum_{t=1}^{T-1}\alpha_t}+\frac{2B^2\sum_{t=1}^{T-1}\alpha_t^2}{m\sum_{t=1}^{T-1}\alpha_t}\\&+ 8 LrB M(T)+ 2\lambda^2B^2M(T)^2,
 \end{align*}
 where $M(T):=\max_{1\leq t\leq T-1}\{\sum_{j=0}^{t-1}\alpha_j\lambda^{t-1-j}\}$ and $r$ is the radius of $V$.
\end{lemma}
It is worth mentioning that the optimization error is established on the average point $\textrm{ave}({\bf x}^T)$ for technical reasons.

\medskip

In the following, we provide the results for the strongly convex setting.
\begin{lemma}\label{th0}
Let $f(\cdot;\xi)$ be $\nu$-strongly convex and Assumptions \ref{ass1}, \ref{ass2}, \ref{ass3} hold, and let $V$ be a closed ball with radius $r>0$, and let $({\bf x}^t)_{1\leq t\leq T}$ be the sequence generated by D-SGD.
When $\alpha_t\equiv\alpha>0$, then
\begin{equation*}
\small
    \begin{aligned}
    &\EE\|{\bf x}^{T}-{\bf x}^*\|^2\leq   (1-2\alpha\nu)^{T-1}\|{\bf x}^{1}-{\bf x}^*\|^2\\
    &+(\frac{4\alpha LrB}{(1-\lambda)\nu} + \frac{\lambda^2B^2\alpha}{m(1-\lambda)^2\nu}) 1_{\lambda\neq 0},
    \end{aligned}
\end{equation*}
where $1_{\lambda\neq 0}=1$ when $\lambda\neq 0$, and $1_{\lambda\neq 0}=0$ when $\lambda= 0$.
When $\alpha_t={1}/{(2\nu(t+1))}$, it then follows
$$\EE\|{\bf x}^{T}-{\bf x}^*\|^2\leq\frac{\|{\bf x}^{1}-{\bf x}^*\|^2}{T-1}+\frac{D_{\lambda}\ln T}{T-1},$$
where $D_{\lambda}:=\frac{B^2}{2\nu^2}+\frac{\lambda^2B^2C_{\lambda}^2}{2\nu^2m}+\frac{2LrBC_{\lambda}}{\nu^2}$ and $$C_{\lambda}:=\left\{
                                                                                                                                           \begin{array}{c}
                                                                                                                                             \ln\frac{1}{\lambda}\frac{\lambda^{\ln\frac{1}{\lambda}}}{\lambda}+\frac{\ln^2\frac{1}{\lambda}}{16\lambda}\lambda^{\frac{\ln\frac{1}{\lambda}}{8}}+\frac{2}{\lambda \ln\frac{1}{\lambda}}~~\lambda\neq 0,  \\
                                                                                                                                             0,~~\lambda=0. \\
                                                                                                                                           \end{array}
\right.$$
\end{lemma}

The result shows that D-SGD with projection converges sublinearly in the strongly convex case. To reach an $\epsilon>0$ error, we shall set the iteration number as $\widetilde{\mathcal{O}}({1}/{\epsilon})$. 
Our result coincides with the existing convergence results of SGD with strong convexity \citep{rakhlin2012making}. What is different is that D-SGD is affected by the parameter $\lambda$, which is determined by the structure of the connected graph\footnote{For the strongly convex case, we avoid showing the result under general step size due to the complicated form.}.

\subsection{General Convexity}
Notice that the computational error of D-SGD, in this case, is described by $\textrm{ave}({\bf x}^T)$. Thus, we need to estimate the generalization bound about $\textrm{ave}({\bf x}^T)$.

\begin{theorem}\label{th4}
Let $f(\cdot;\xi)$ be convex and  Assumptions \ref{ass1}, \ref{ass2}, \ref{ass3} hold. If the step size $\alpha_t\equiv\alpha\leq {2}/{L}$, then the average  output \eqref{average} obeys the following generalization  bound
\begin{equation*}
\footnotesize
    \begin{aligned}
    &\epsilon_{\emph{ex-gen}} \leq \frac{2B^2\alpha(t-1)}{mn}+\frac{4\alpha B^2( 1+\alpha B)(t-1)}{1-\lambda}1_{\lambda\neq 1}\\
    &+\frac{4r^2}{(T-1)\alpha}+\frac{2B^2\alpha}{m}+ \frac{8 LrB\alpha}{1-\lambda}1_{\lambda\neq 1}+ \frac{2\lambda^2B^2\alpha^2}{(1-\lambda)^2}.
    \end{aligned}
\end{equation*}
Furthermore, if the step size is chosen as $\alpha_t={1}/{(t+1)}$, we have
\begin{equation*}
\footnotesize
    \begin{aligned}
    &\epsilon_{\emph{ex-gen}}\leq \frac{B^2\ln T}{mn}+\frac{4 B^2( 1+ B)}{\ln (T+1)}1_{\lambda\neq 1}+  2\lambda^2B^2C_{\lambda}^2\\
    &+\frac{4r^2}{\ln (T+1)}+\frac{4B^2}{m\ln (T+1)}+  8 LrBC_{\lambda}1_{\lambda\neq 1}.
    \end{aligned}
\end{equation*}
\end{theorem}

  \begin{figure}[t!]
    \centering
  \subfigure[Random ]{ \includegraphics[width=0.18\textwidth]{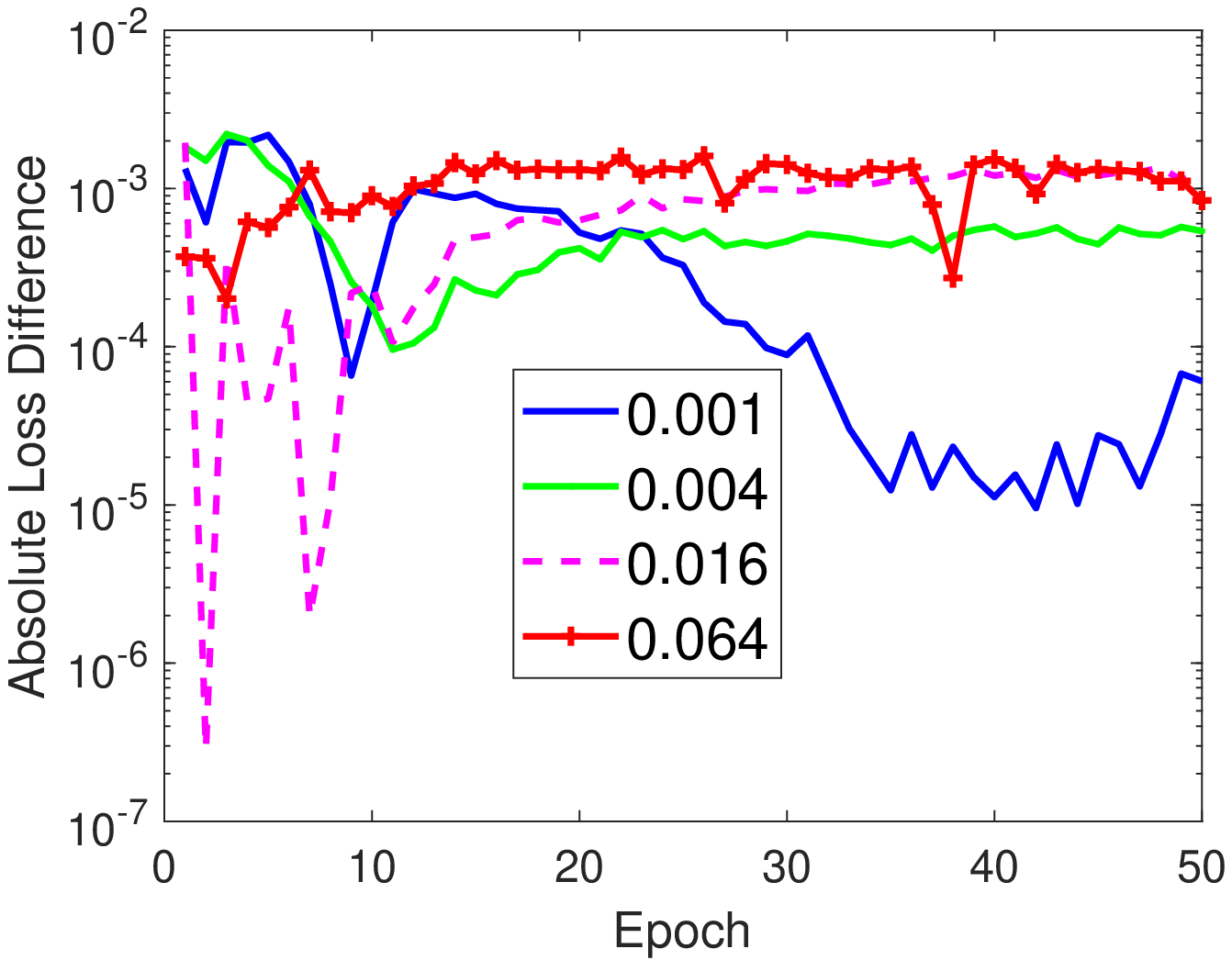}}
  \subfigure[Star ]{\includegraphics[width=0.18\textwidth]{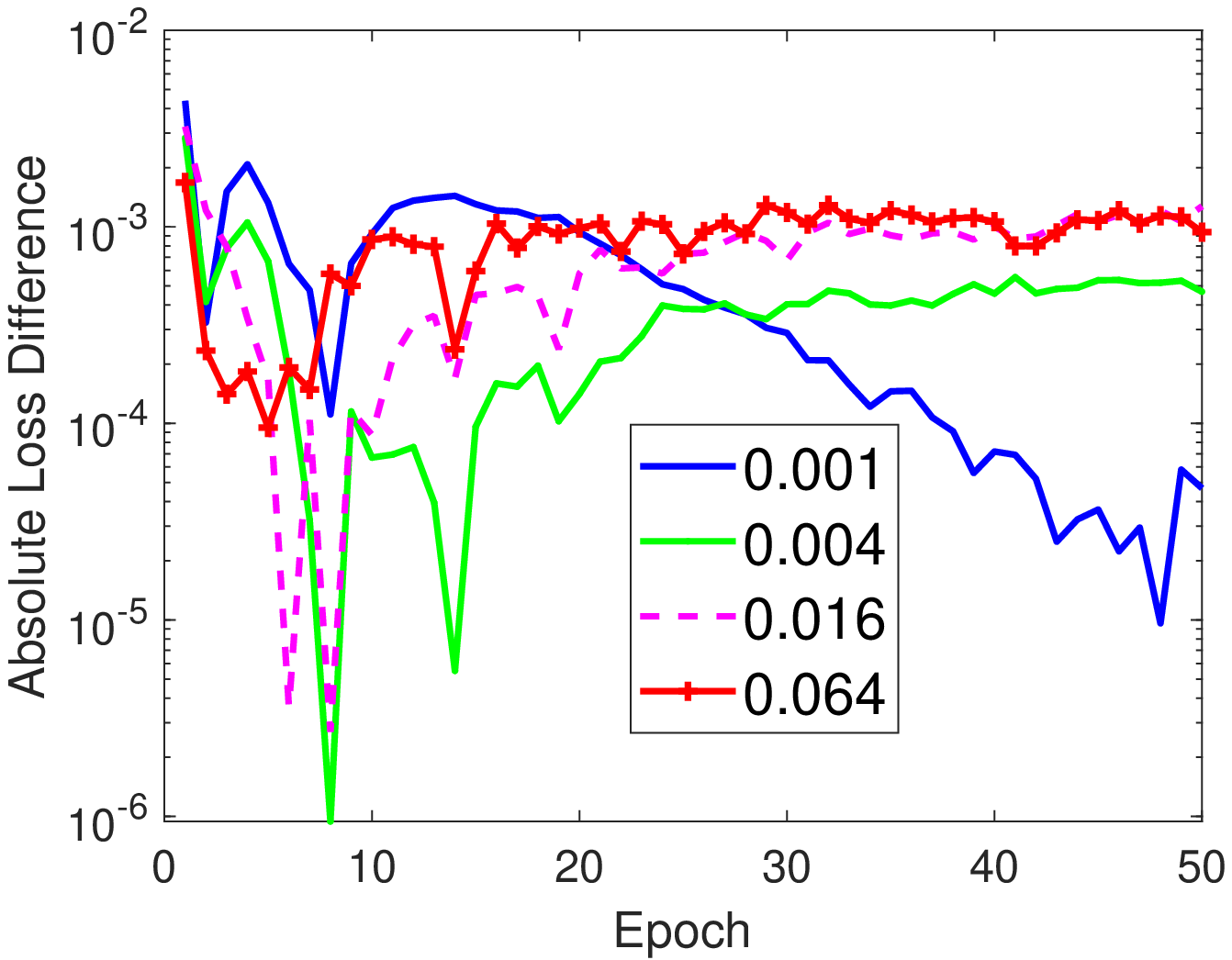}}
  \subfigure[Cycle]{\includegraphics[width=0.18\textwidth]{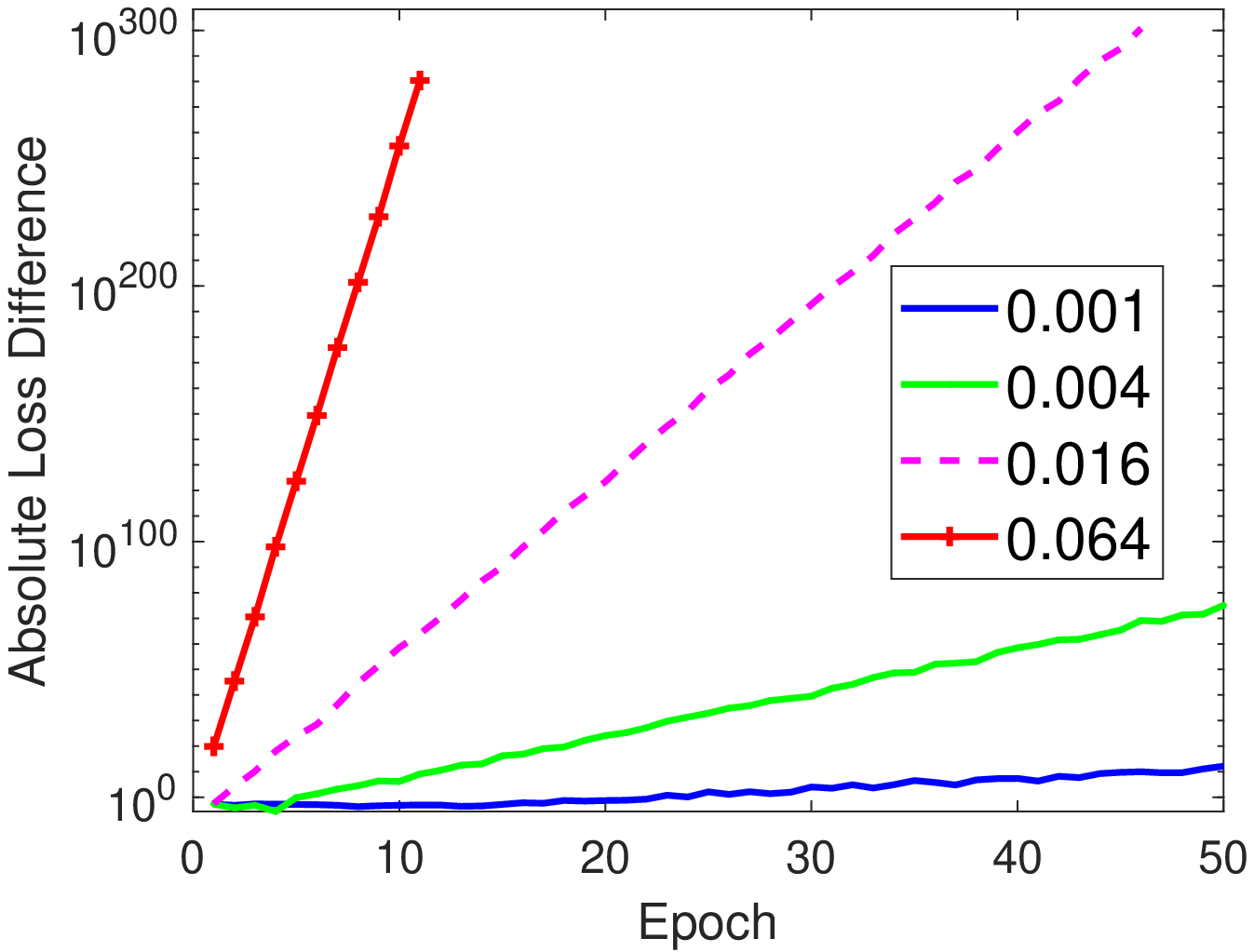}}
  \subfigure[k-NNG]{\includegraphics[width=0.18\textwidth]{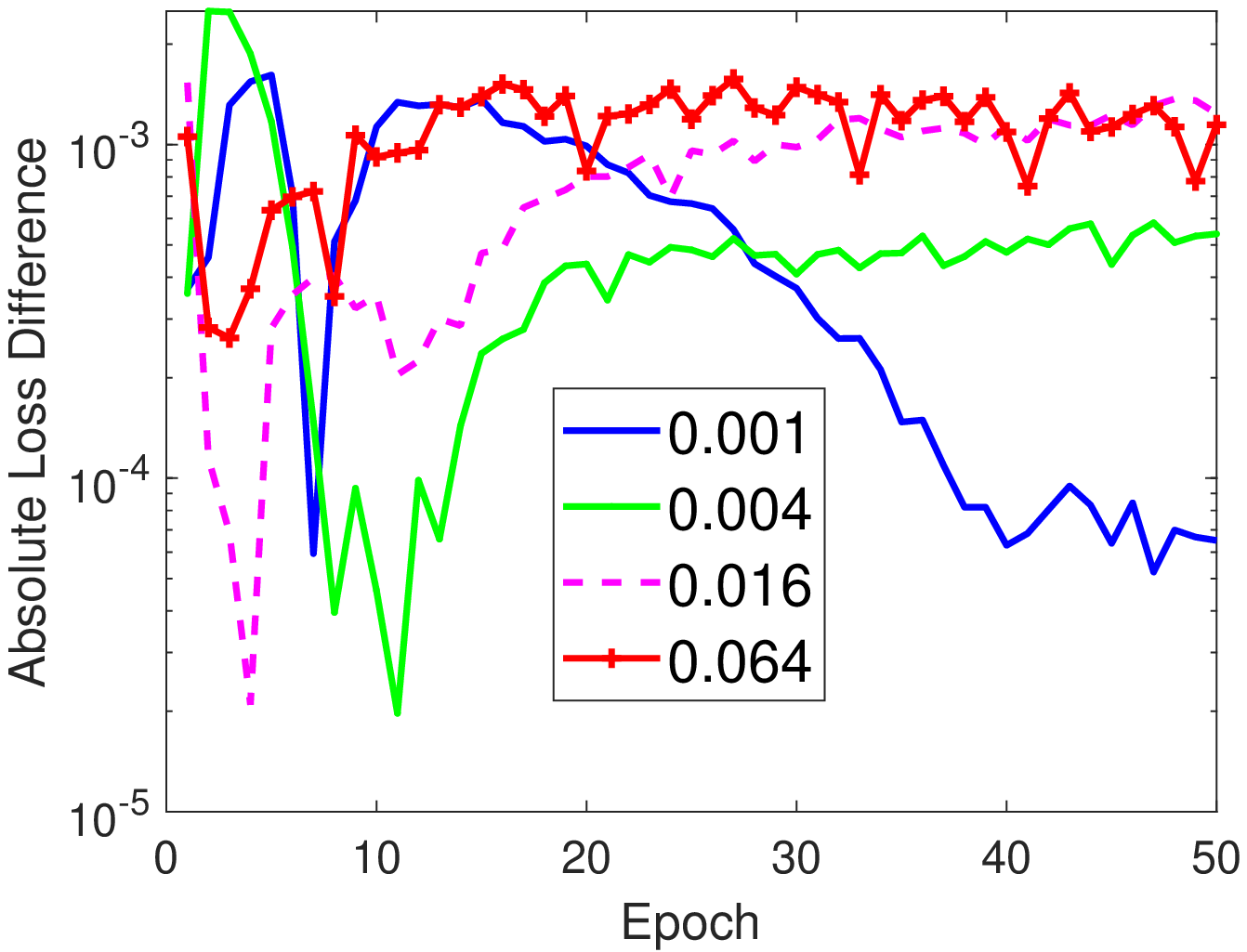}}
      \subfigure[Bipartite]{ \includegraphics[width=0.18\textwidth]{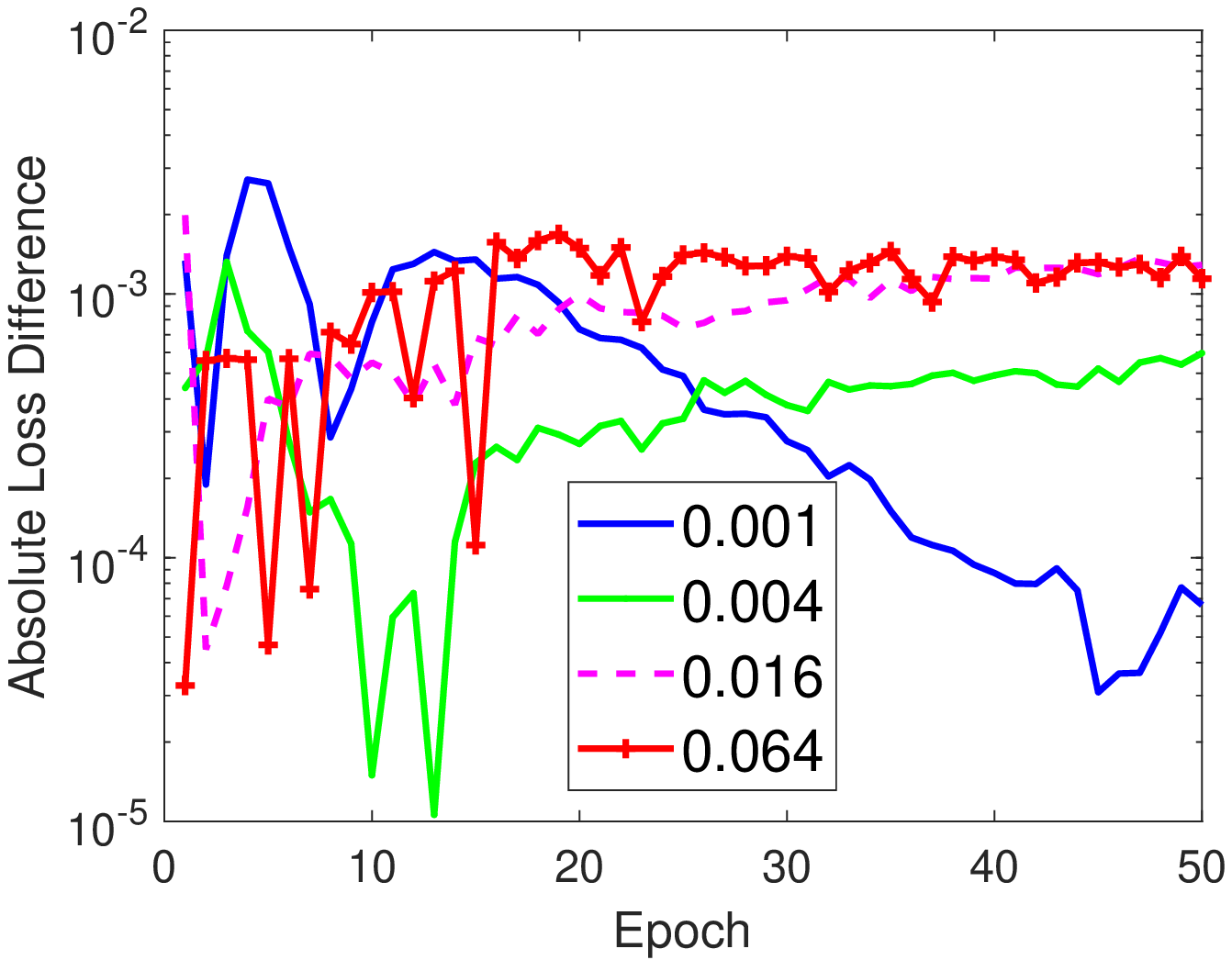}}
        \subfigure[Complete]{ \includegraphics[width=0.18\textwidth]{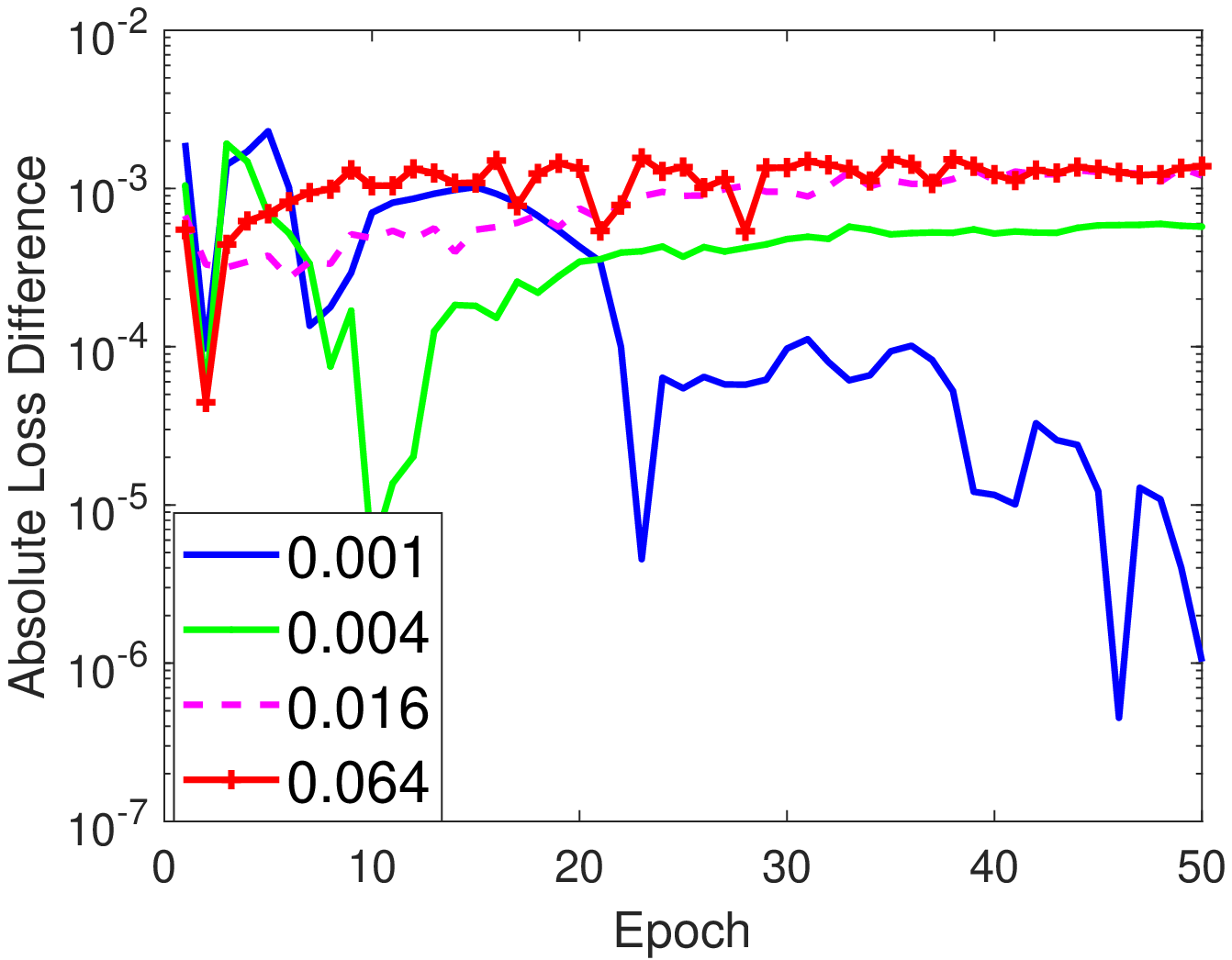}}
    \caption{Absolute loss difference versus epochs for linear regression task on Body Fat dataset. With current learning rates, D-SGD diverges for the Cycle graph. With enough iterations, the smaller learning rate can achieves a smaller difference for all graph. }
    \label{convex}
\end{figure}
  \begin{figure}[t!]
    \centering
  \subfigure[Random ]{ \includegraphics[width=0.18\textwidth]{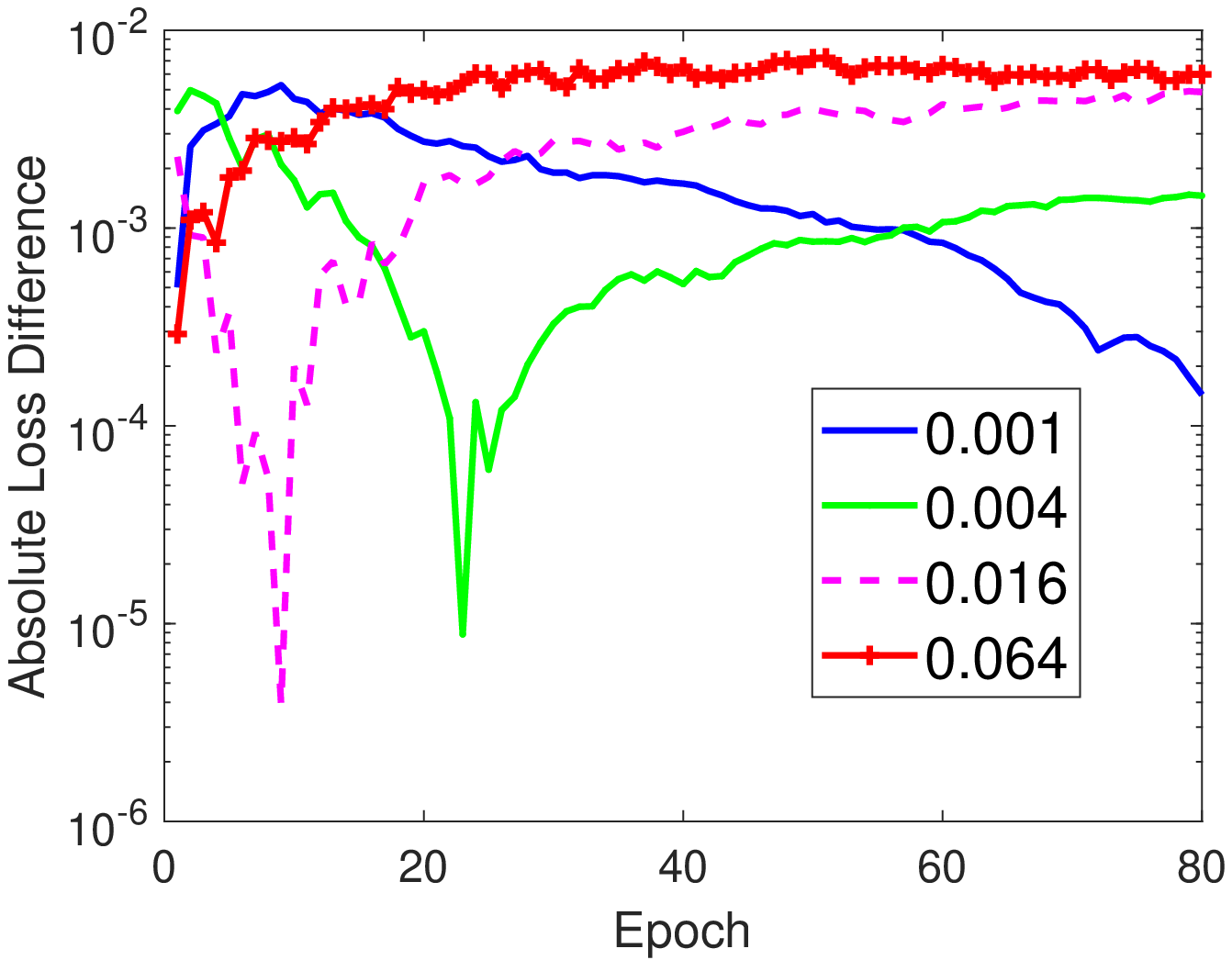}}
  \subfigure[Star ]{\includegraphics[width=0.18\textwidth]{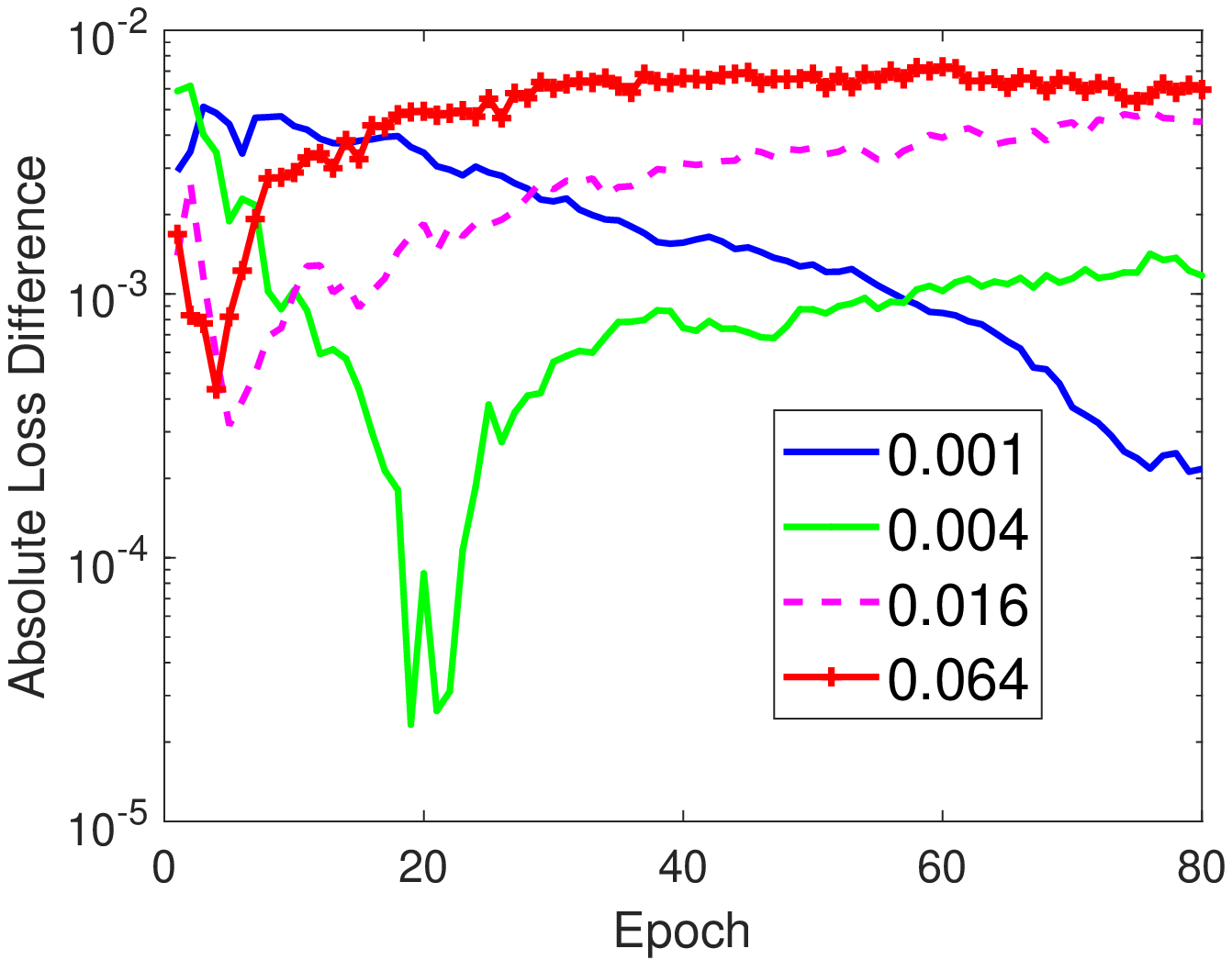}}
  \subfigure[Cycle]{\includegraphics[width=0.18\textwidth]{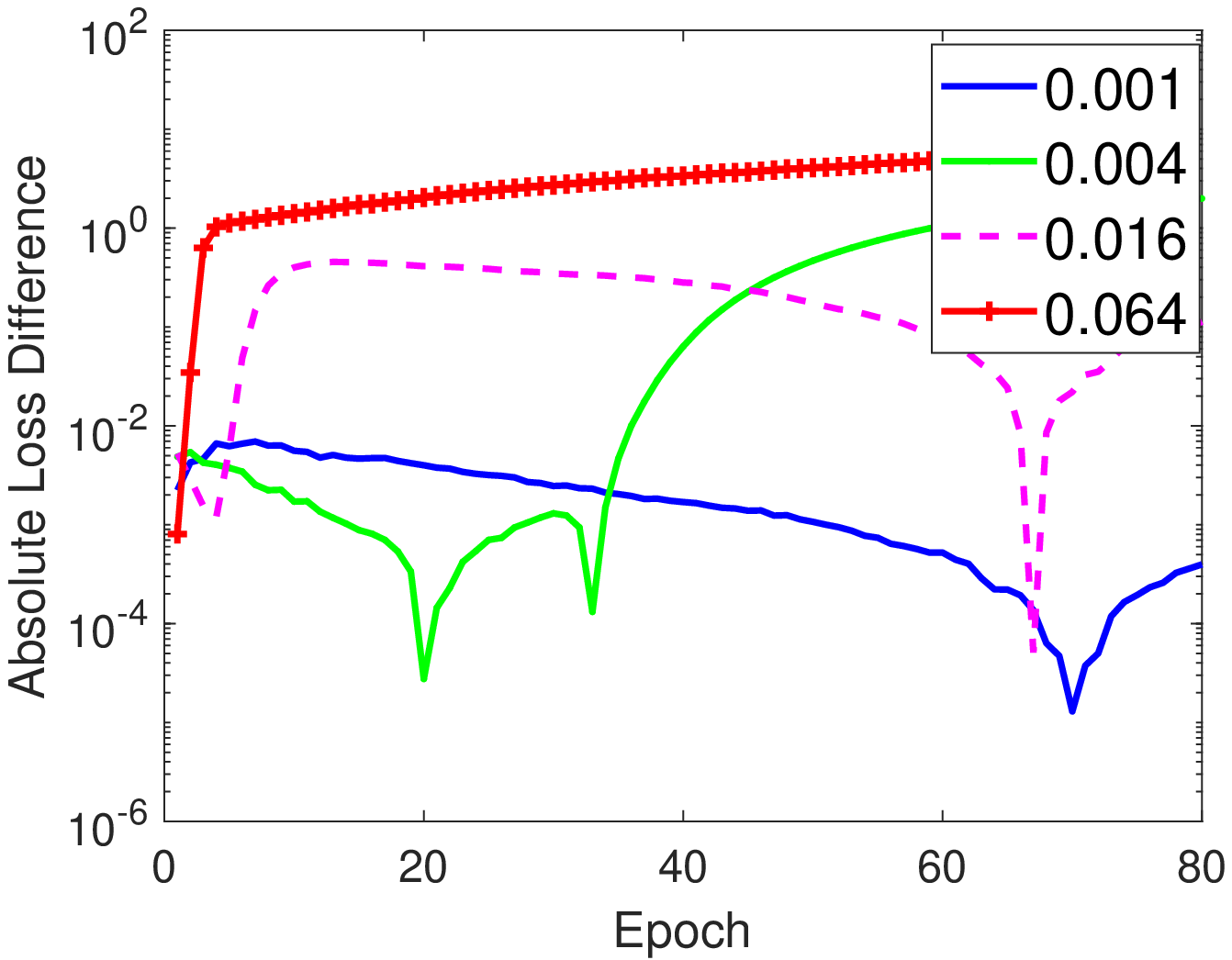}}
  \subfigure[k-NNG]{\includegraphics[width=0.18\textwidth]{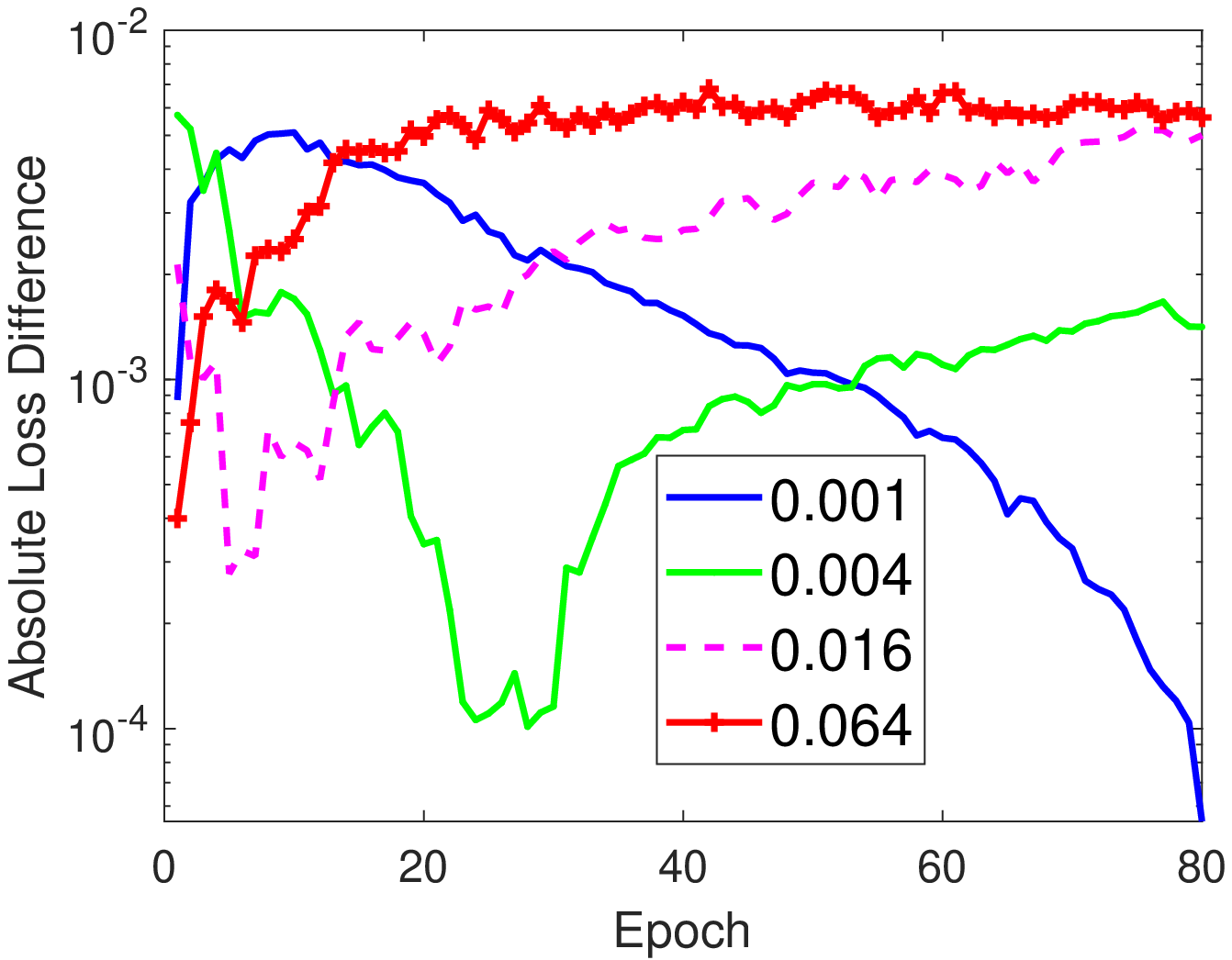}}
      \subfigure[Bipartite]{ \includegraphics[width=0.18\textwidth]{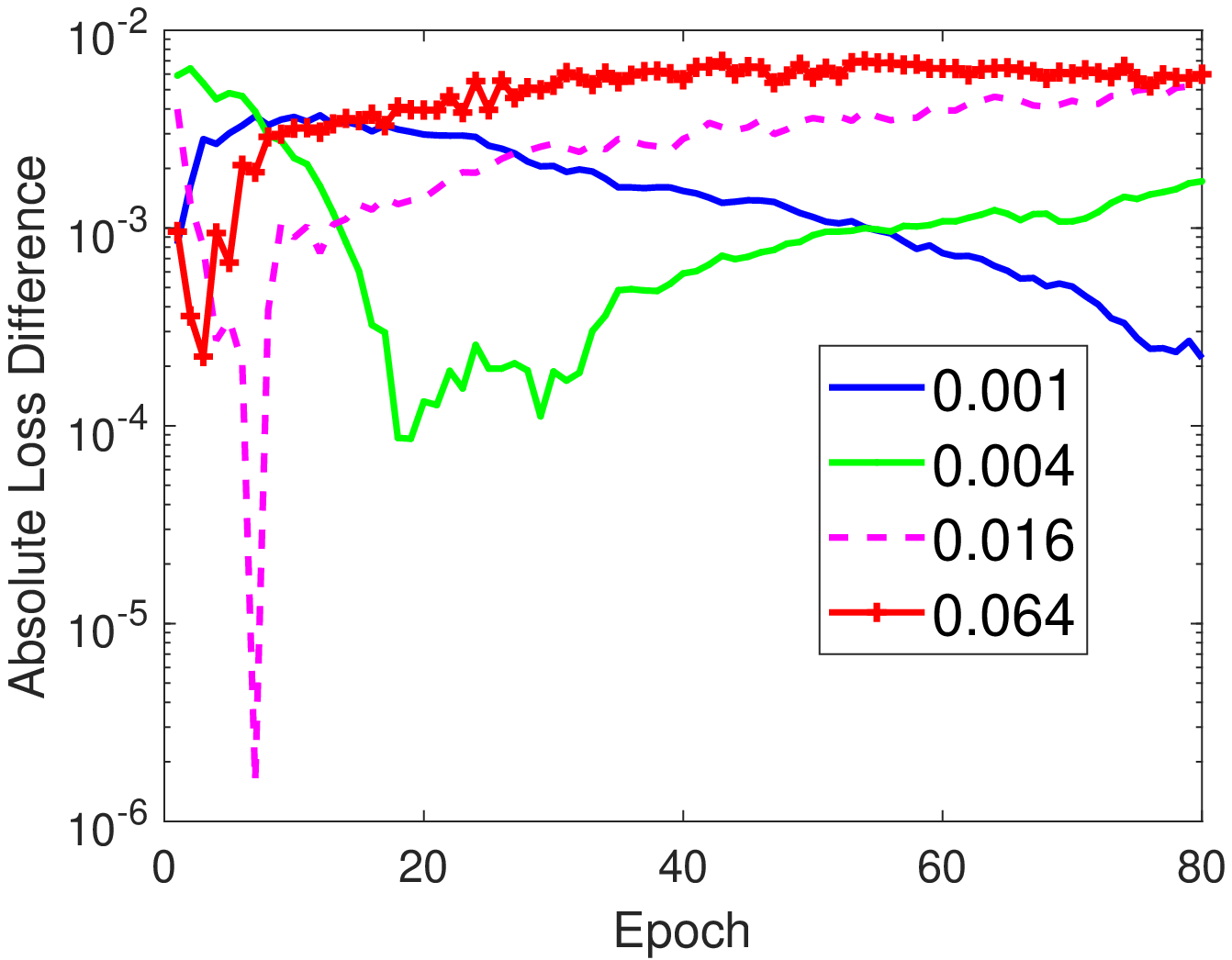}}
        \subfigure[Complete]{ \includegraphics[width=0.18\textwidth]{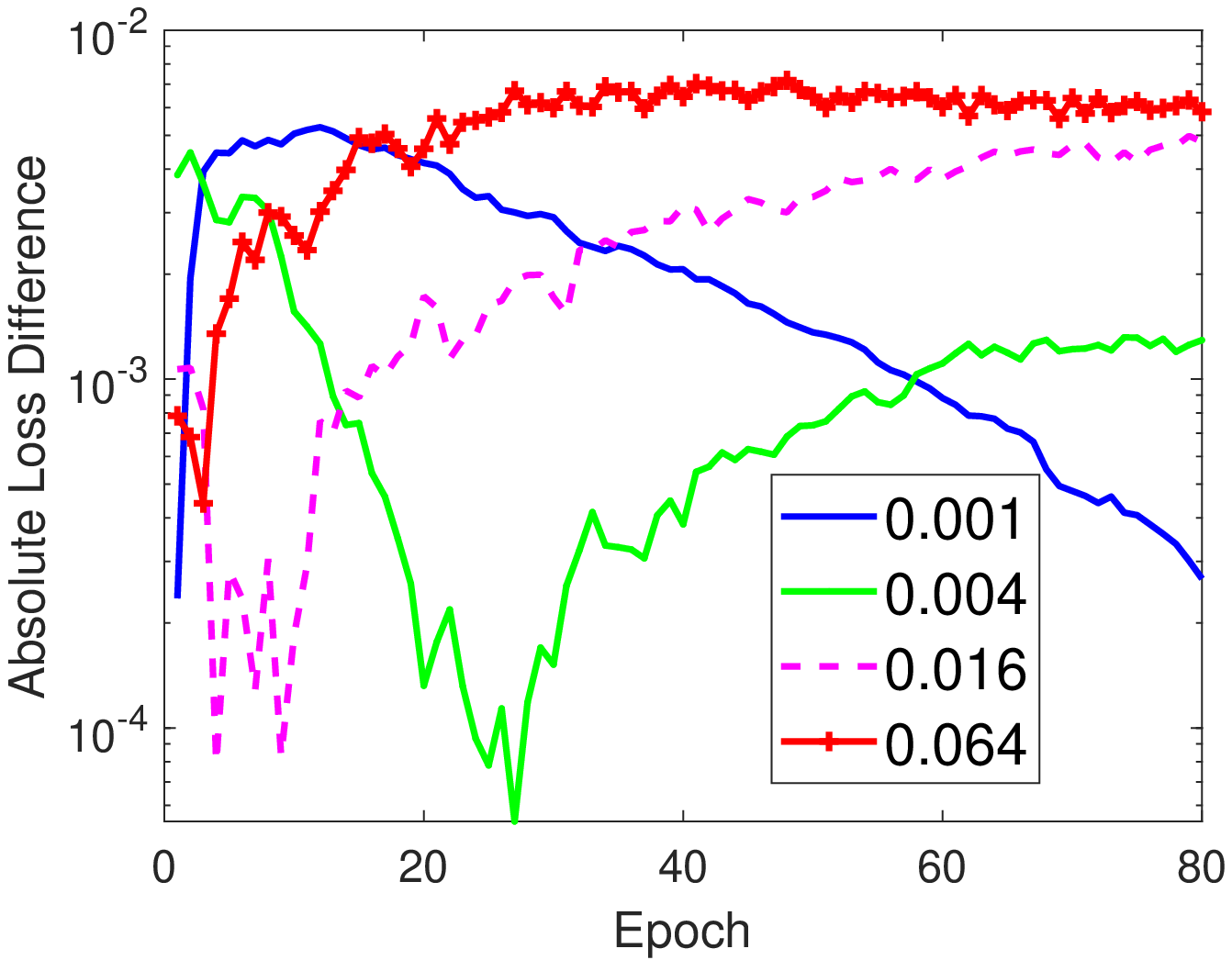}}
    \caption{ Absolute loss difference versus epochs for the $\ell_2$ regularized logistic regression task on ijcnn1.
    The strongly convex tests are similar to general convex ones but more smooth. With strong convexity, D-SGD converges over the Cycle graph. In particular, when a smaller learning rate is used. }
    \label{sconvex}
\end{figure}

  \begin{figure}[t!]
    \centering
  \subfigure[Random ]{ \includegraphics[width=0.18\textwidth]{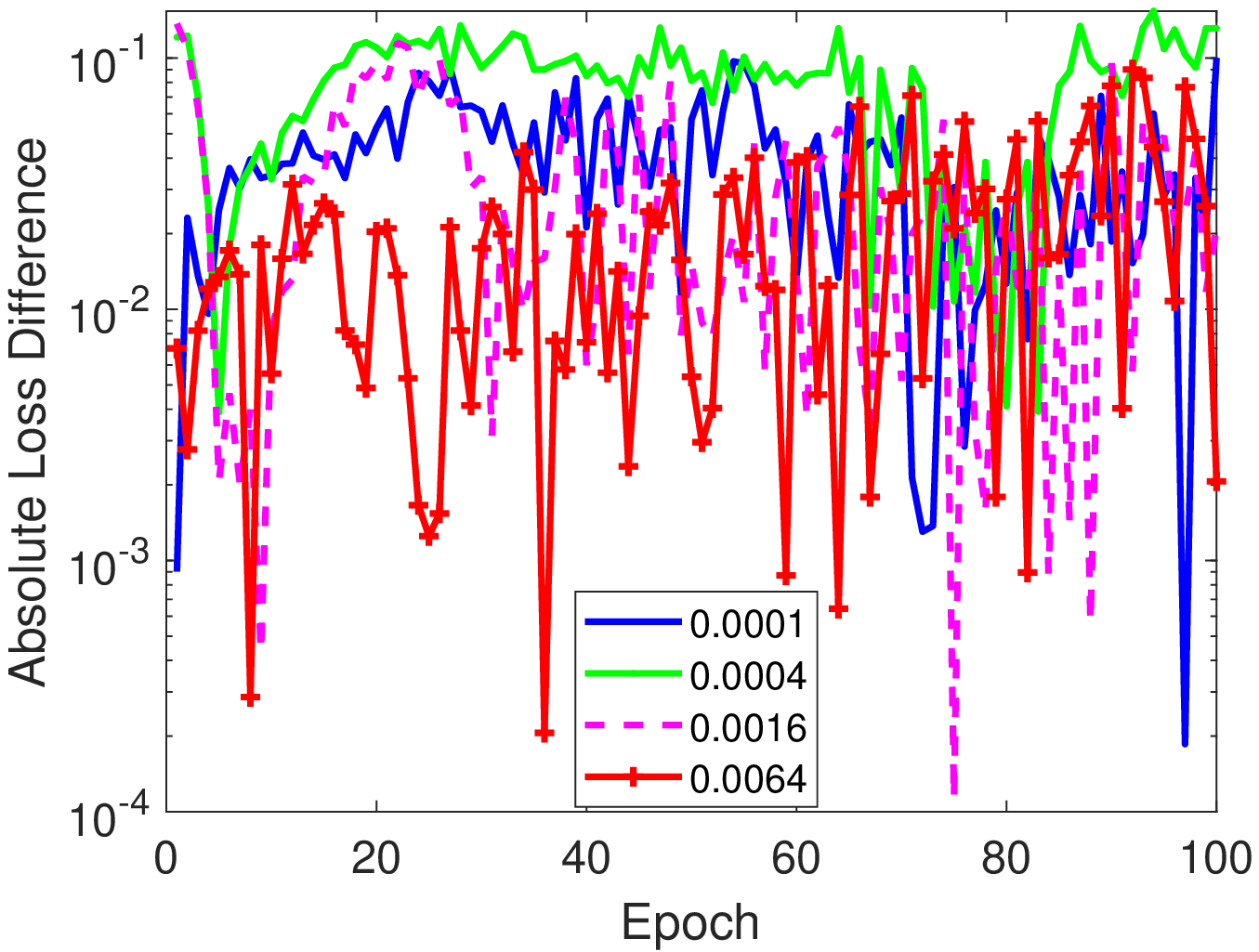}}
  \subfigure[Star ]{\includegraphics[width=0.18\textwidth]{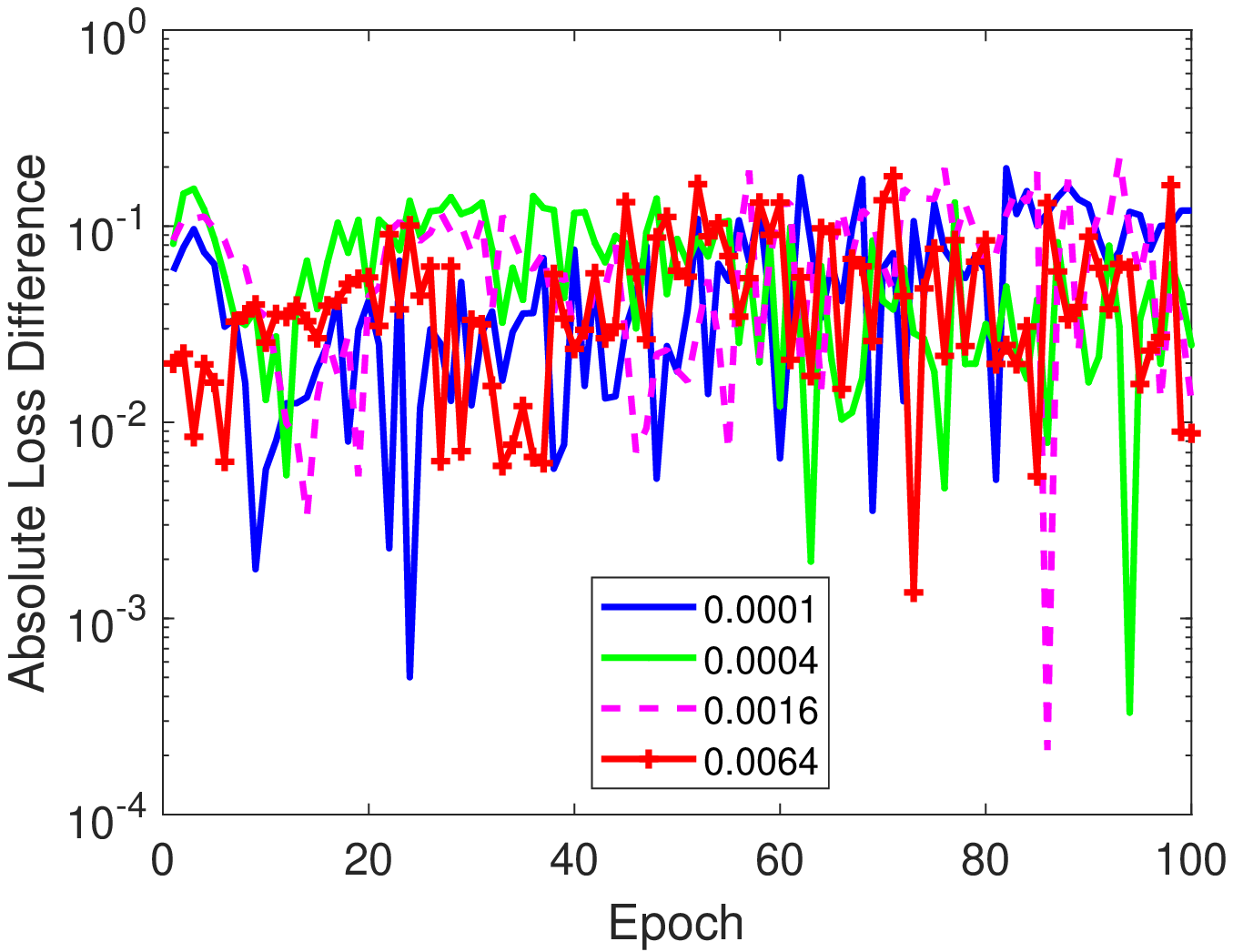}}
  \subfigure[Cycle]{\includegraphics[width=0.18\textwidth]{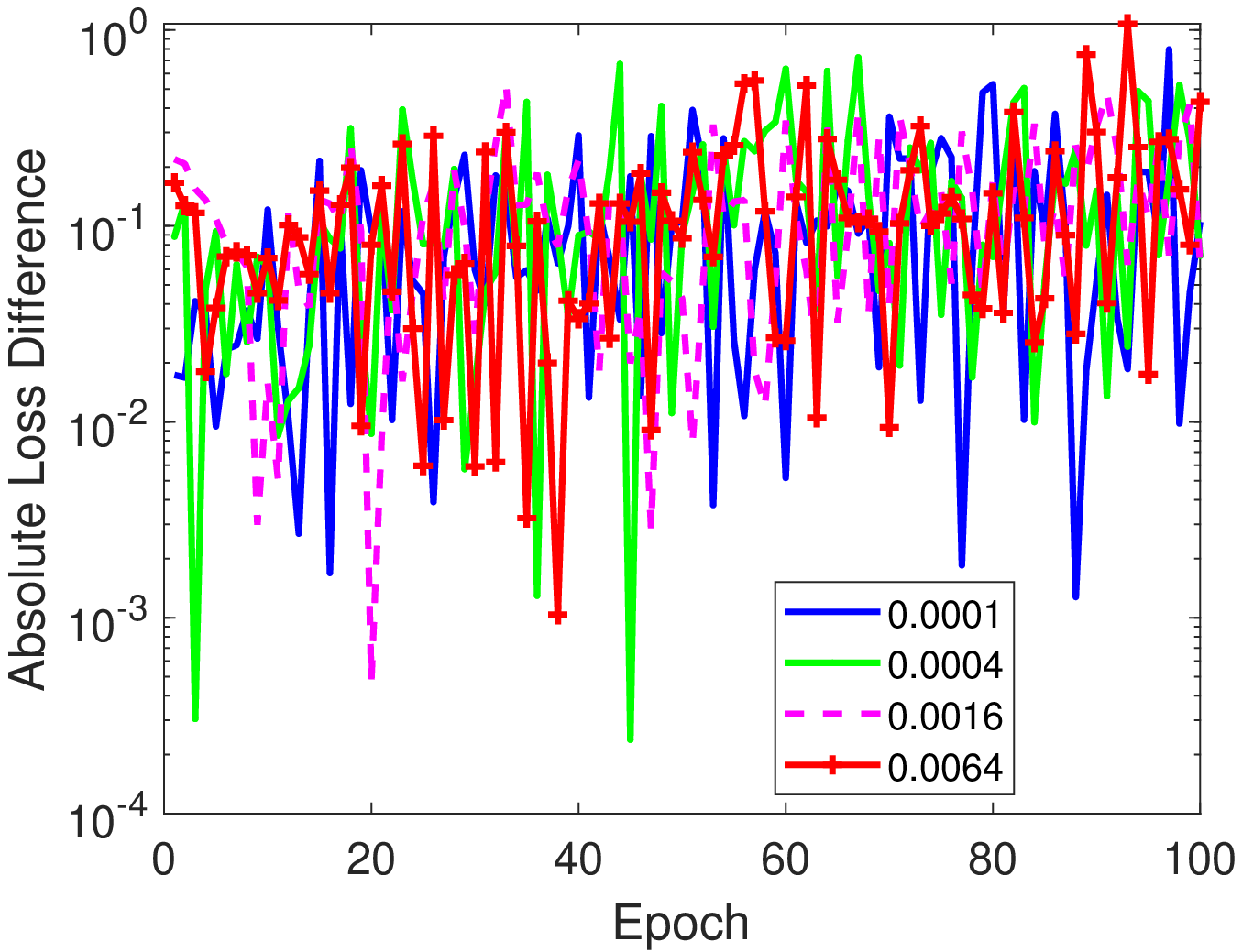}}
  \subfigure[k-NNG]{\includegraphics[width=0.18\textwidth]{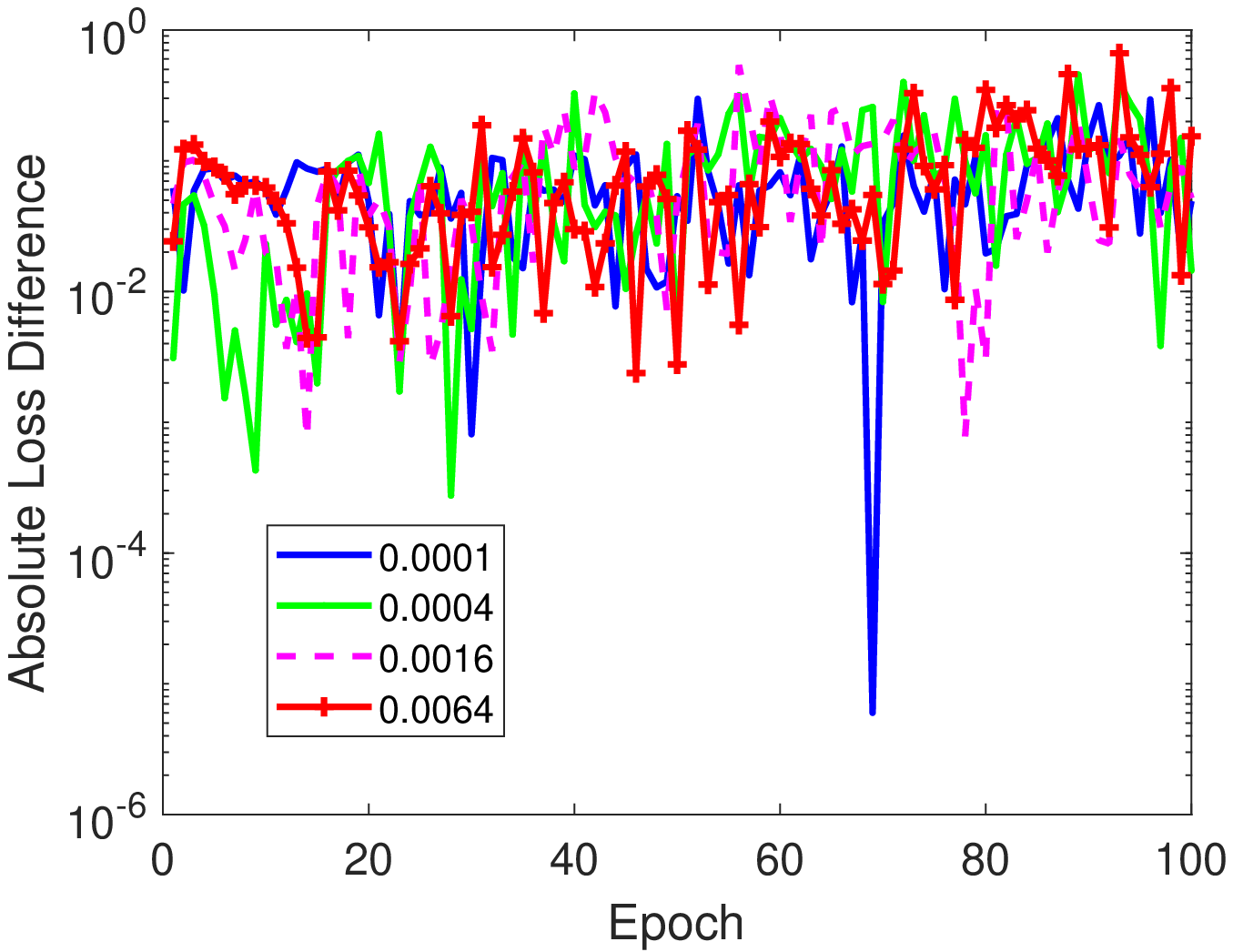}}
      \subfigure[Bipartite]{ \includegraphics[width=0.18\textwidth]{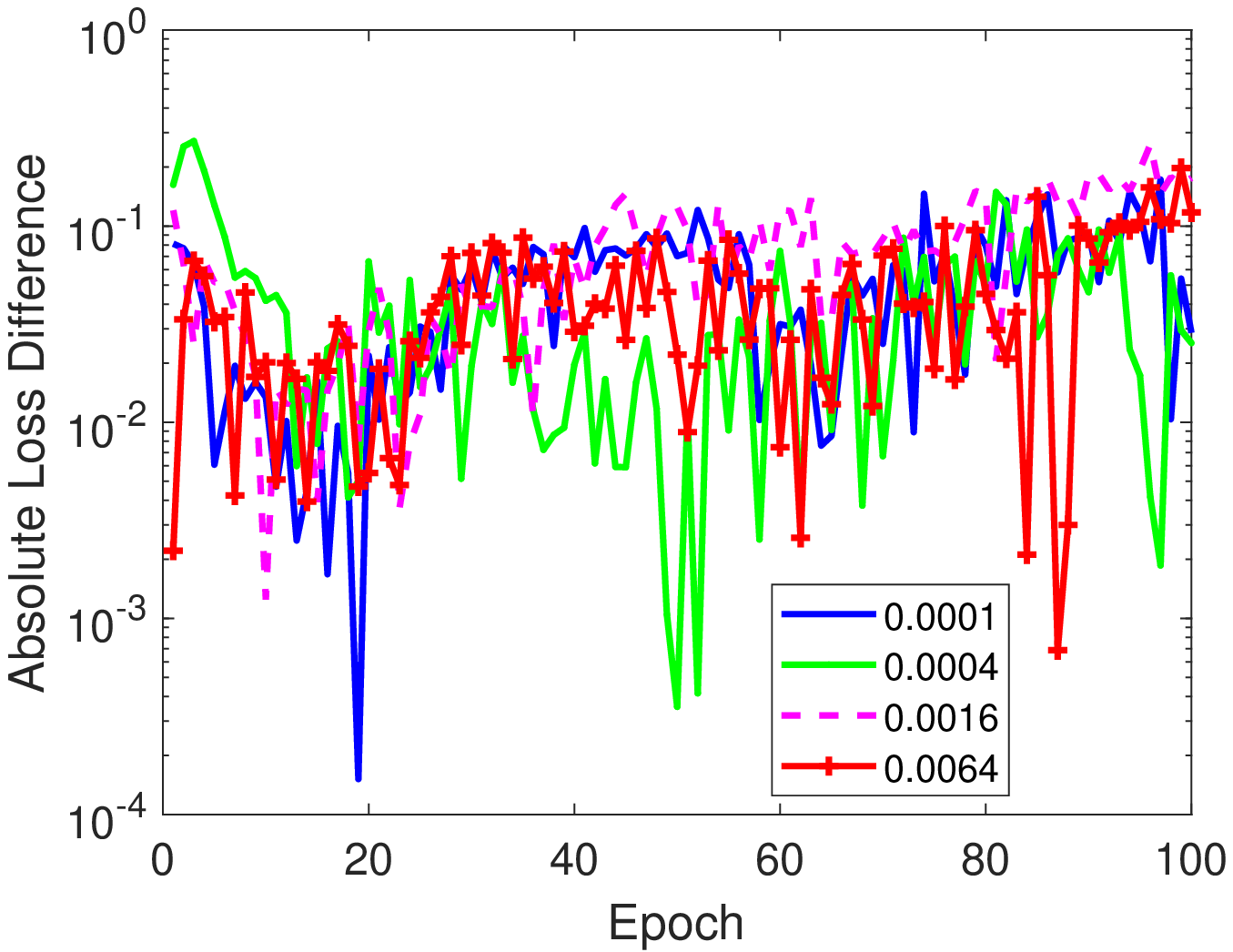}}
        \subfigure[Complete]{ \includegraphics[width=0.18\textwidth]{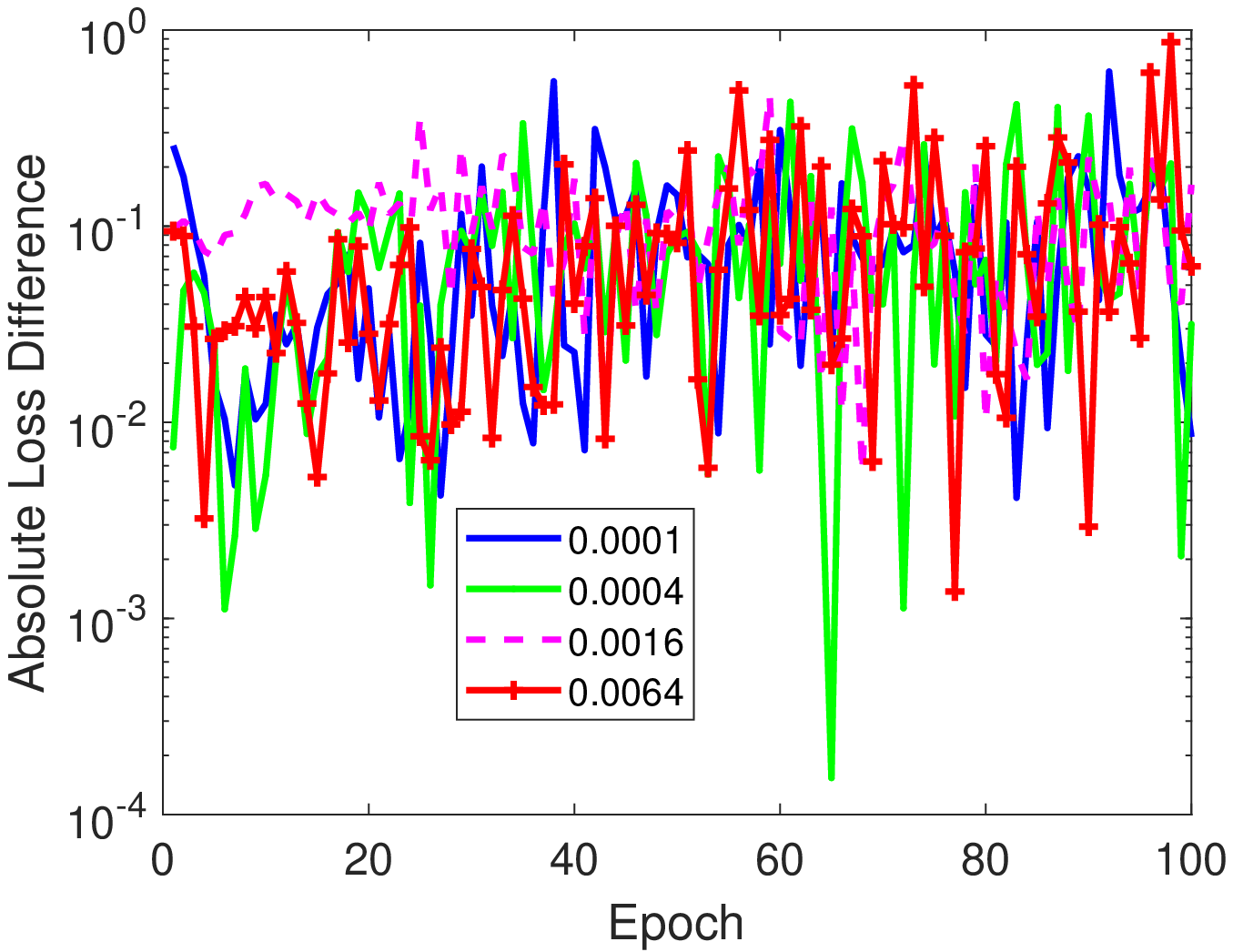}}
    \caption{ Absolute loss difference versus epochs for training nonconvex machine learning model, i.e., ResNet20. Unlike the convex cases, the absolute loss difference oscillates chaotically, which implies worse stability.
    }
    \label{nonconvex}
\end{figure}
\subsection{Strong Convexity}
Now, we present the excess generalization of D-SGD under strong convexity.
\begin{theorem}\label{th5}
Let $f(\cdot;\xi)$ be $\nu$-strongly convex and  Assumptions \ref{ass1}, \ref{ass2}, \ref{ass3} hold. If the step size $\alpha_t\equiv\alpha\leq {1}/{L}$, the excess generalization  bound is
\begin{equation*}
\footnotesize
    \begin{aligned}
    &\epsilon_{\emph{ex-gen}}\leq  \frac{2B^2}{mn\nu}+\frac{4( 1+\alpha B)B^2}{\nu} \frac{1_{\lambda\neq 0}}{1-\lambda}\\
    &+B\sqrt{(1-2\alpha\nu)^{T-1}4r^2+(\frac{4\alpha LrB}{(1-\lambda)\nu} + \frac{\lambda^2B^2\alpha}{m(1-\lambda)^2\nu}) 1_{\lambda\neq 0}}.
    \end{aligned}
\end{equation*}
Furthermore, if the step size $\alpha_t={1}/{(\nu (t+1))}$, the excess generalization  bound is
\begin{equation*}
\footnotesize
    \begin{aligned}
    \epsilon_{\emph{ex-gen}}&\leq  \frac{2B^2}{mn\nu}+\frac{4( \nu+ B)B^2}{\nu^2} \frac{1_{\lambda\neq 0}}{1-\lambda}+B\sqrt{\frac{4r^2}{T-1} +\frac{D_{\lambda}\ln T}{T-1}}.
    \end{aligned}
\end{equation*}
\end{theorem}

\section{Numerical Results}
We numerically verify our theoretical findings in this section, with a focus on testing three kinds of models, namely, strongly convex, convex, and nonconvex.
For all the above three scenarios, we set the number of nodes $m$ to 10 and conduct two kinds of experiments: the first kind of experiments is to verify the stability and generalization results. Given a fixed graph, we use two sets of samples that are of the same amount, and the entries are differing by a small portion. We compare the training loss and training accuracy of D-SGD on these two datasets; the second kind is to demonstrate the effects due to the structure of the connected graph. We run our experiments on different types of connected graphs with the same dataset. In particular, we test six different connected graphs, as shown in Figure~\ref{graphs}.


\subsection{Convex case}
We consider the following optimization problem
  \begin{equation*}
    \min_{{\bf x}\in \RR^{14}}  \Phi({\bf x}):=\frac{1}{504}\sum_{i=1}^{252}\|\xi_i^{\top}{\bf x}- {\bf y}_i\|^2,
  \end{equation*}
which arises from a simple regression problem. Here, we use the Body Fat dataset \cite{johnson1996fitting} which contains 252 samples. We run D-SGD on two subsets of the Body Fat dataset, and both of size 200. Let ${\bf x}^k$  and $\widehat{{\bf x}}^k$ be the outputs of the D-SGD on the two different subsets. We define the absolute loss difference as
$|\Phi({\bf x}^k)- \Phi(\widehat{{\bf x}}^k)|.$
For the above six graphs, we record the absolute difference in the value of function $\Phi$ for 
a set of learning rate, namely,
$\{0.001,0.004,0.016,0.064\}$ in Figure \ref{convex}. In the second test, we use  the learning rate $0.001$ and compare the absolute loss difference with different graphs in Figure \ref{comparison} (a).
Our results show that the smaller learning rate usually yields a smaller loss difference, and the complete graph can achieve the smallest bound. These observations are consistent with our theoretical results for the convex D-SGD.

\subsection{Strongly convex case}
To verify our theory on the strongly convex case, we consider the regularized logistic regression model as follows
\begin{equation*}
\small
\min_{{\bf x}\in\mathbb{R}^{22}}\left\{\frac{1}{9000}\sum_{i=1}^{9000}\left(\log(1+\textrm{exp}(-b_i{\bf a}^{\top}_i{\bf x})) +\frac{\lambda}{2}\|{\bf x}\|^2\right)\right\}.
\end{equation*}
We use the benchmark ijcnn1 dataset\citep{rennie2001improving} and set  $\lambda=10^{-4}$. Two 8000-sample sub-datasets with 1000 different samples are used as the test set. We conduct experiments on the two datasets with the same set of learning rates that are used in the last subsection.
The absolute loss difference under different learning rates is plotted in Figure \ref{sconvex}, and the performance under different graphs is reported in Figure \ref{comparison} (b). The results of D-SGD in the strongly convex case is similar to the convex case. Also, note that the absolute loss difference increases as the learning rates grow.

\subsection{Nonconvex case}
We test ResNet-20 \citep{he2016deep} for CIFAR10 classification \cite{Krizhevsky2009learning}.
We adopt two different 40000-sample subsets. The loss values are built on the test set.
The absolute loss difference with the learning rate set $\{0.0001,0.0004,0.0016,0.0064\}$  versus the epochs is presented in Figure \ref{nonconvex}, and the absolute loss difference with different graphs are shown in Figure \ref{nonconvex} (c). 100 epochs are used in the nonconvex test. The results show that the nonconvex D-SGD is much more unstable than the convex ones, which matches our theoretical findings.

  \section{Conclusion}
In this paper, we develop the stability and generalization error for the (projected) decentralized stochastic gradient descent (D-SGD) in strongly convex, convex, and nonconvex settings. In contrast to the previous works on the analysis of the projected decentralized gradient descent, our theories are built on much more relaxed assumptions. Our theoretical results show that the stability and generalization of D-SGD depend on the learning rate and the structure of the connected graph. Furthermore, we prove that decentralization deteriorates the stability of D-SGD. Our theoretical results are empirically supported by experiments on  training different machine learning models in different decentralization settings. There are numerous avenues for future work: 1) deriving the improved stability and generalization bounds of D-SGD in the general convex and nonconvex cases, 2) proving the high probability bounds, 3) studying the stability and generalization bound of the moment variance of D-SGD.



\onecolumn
\appendix
\begin{center}
{\Large \bf Supplementary materials for}
\end{center}

\begin{center}
\large \bf \textit{Stability and Generalization  Bounds of Decentralized Stochastic Gradient Descent}
\end{center}
\section{More results of the test}
We present the absolute accuracy difference in the decentralized neutral  networks training.
  \begin{figure}[!htb]
    \centering
  \subfigure[Random ]{ \includegraphics[width=0.4\textwidth]{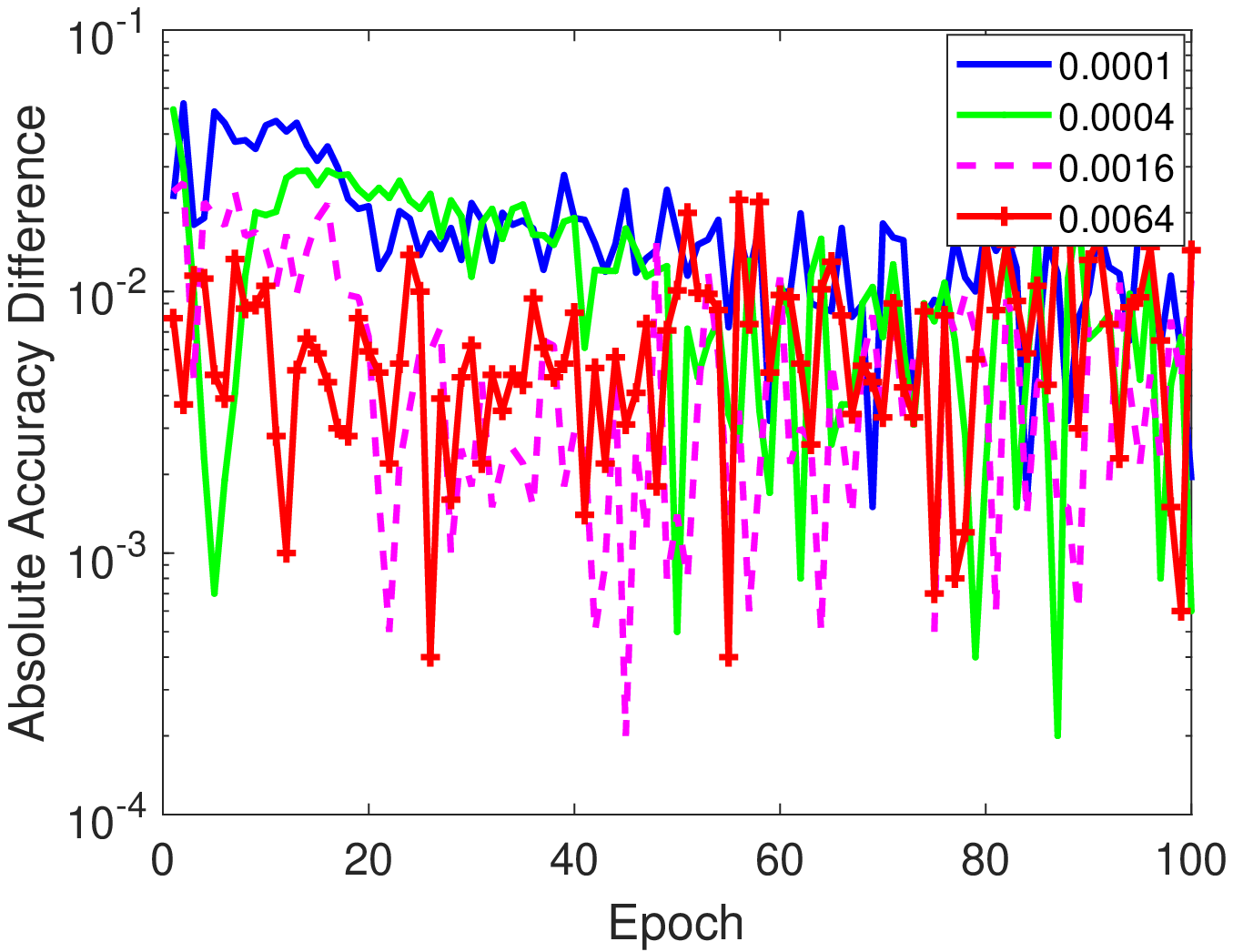}}
  \subfigure[Star ]{\includegraphics[width=0.4\textwidth]{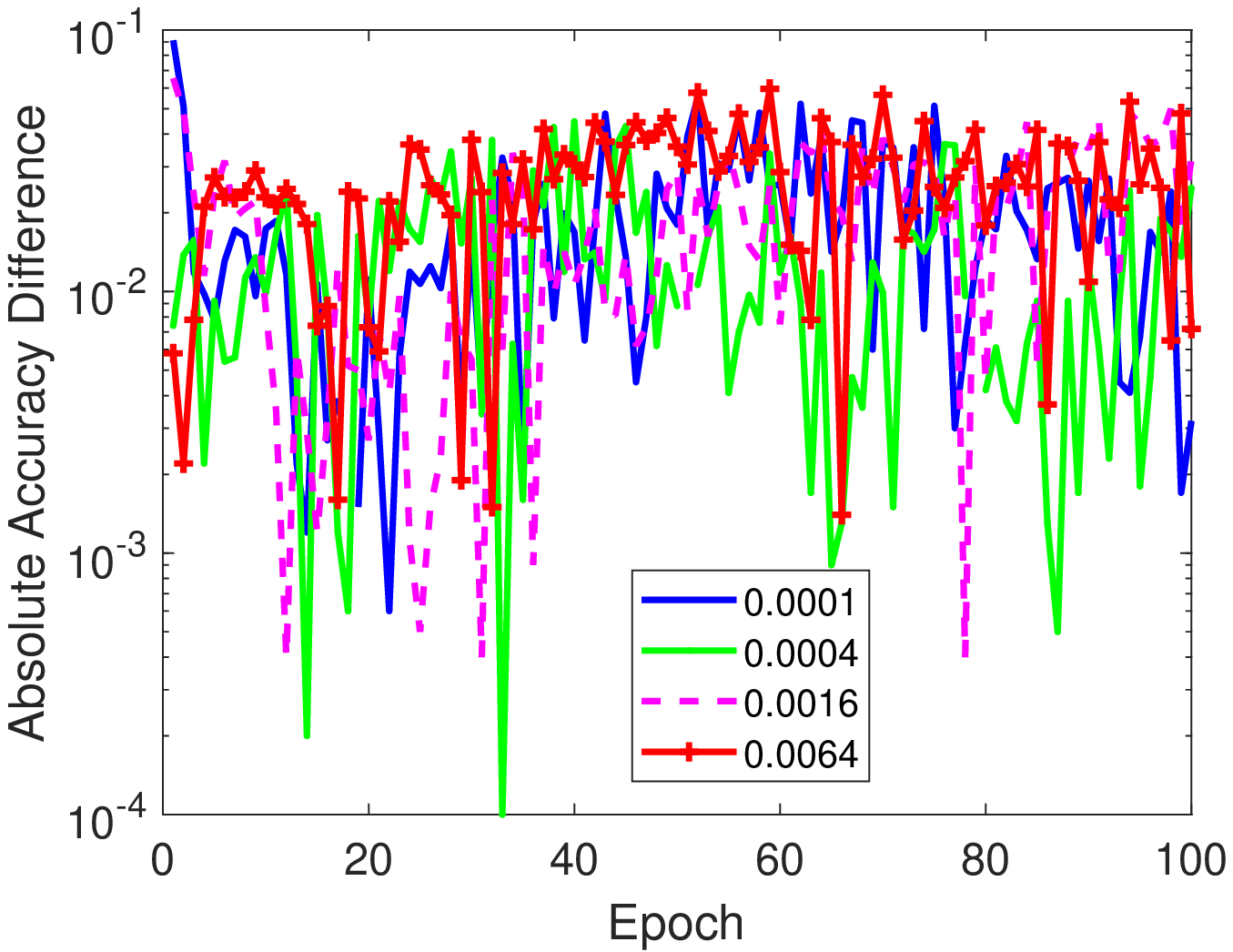}}
  \subfigure[Cycle]{\includegraphics[width=0.4\textwidth]{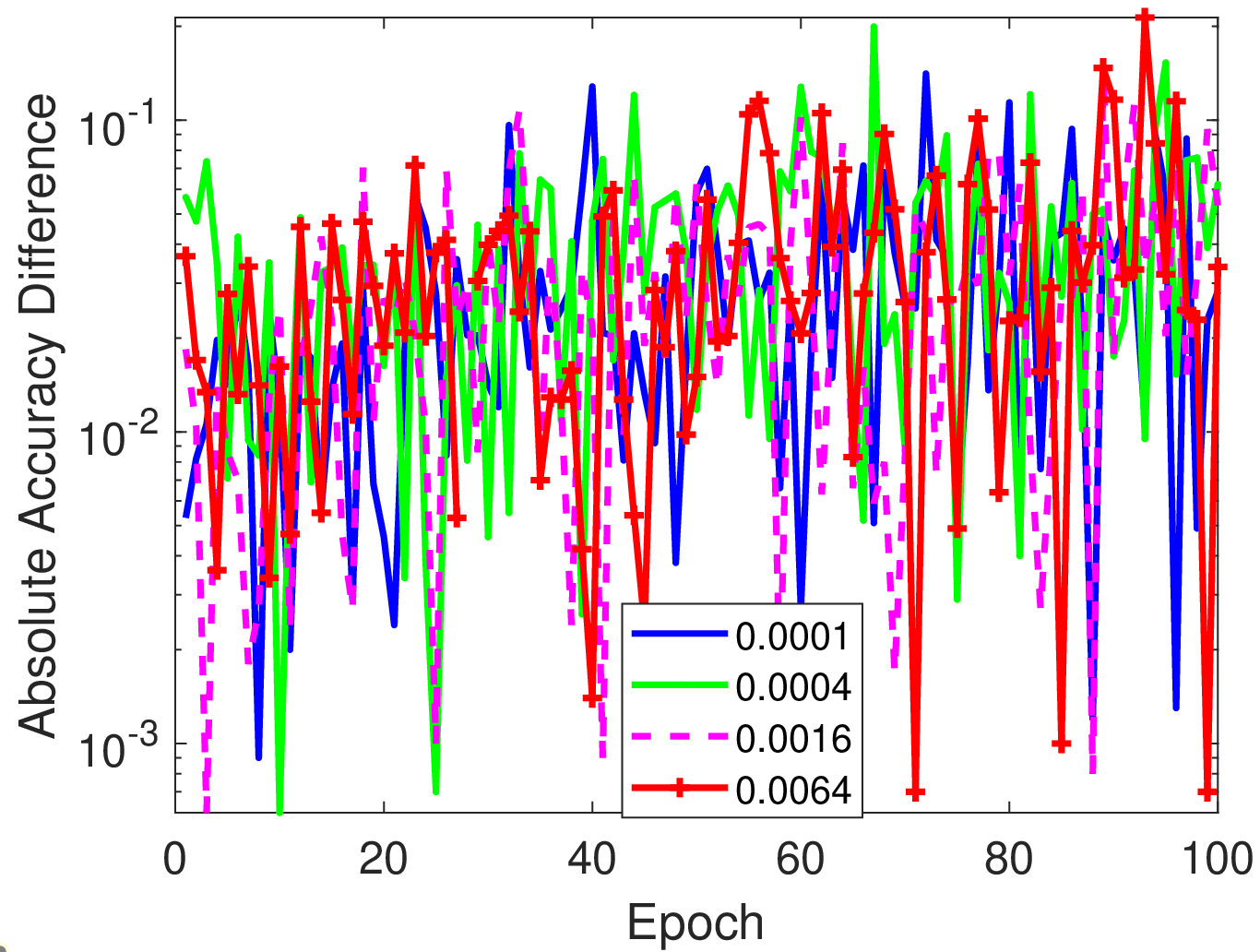}}
  \subfigure[k-NNG]{\includegraphics[width=0.4\textwidth]{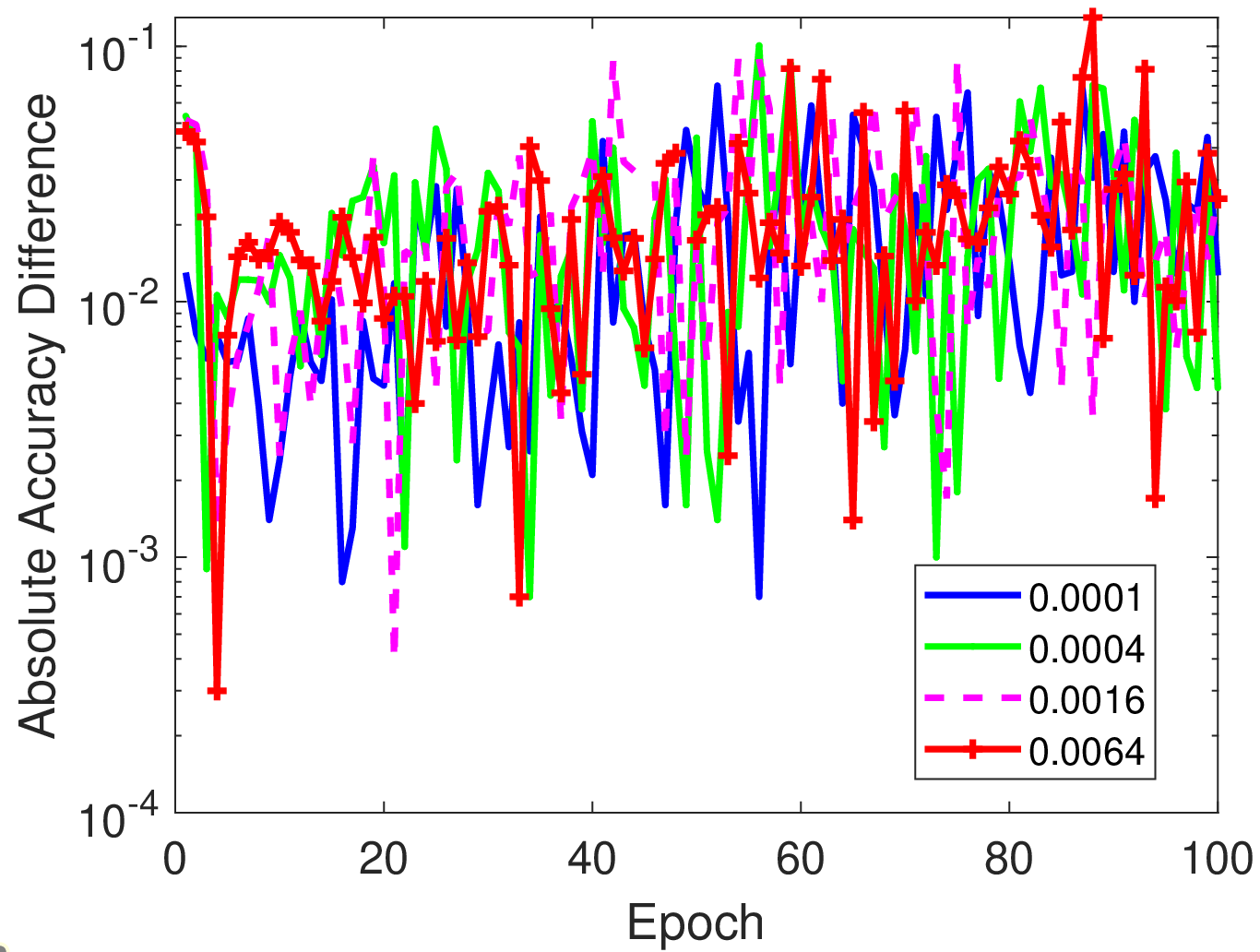}}
      \subfigure[Bipartite]{ \includegraphics[width=0.4\textwidth]{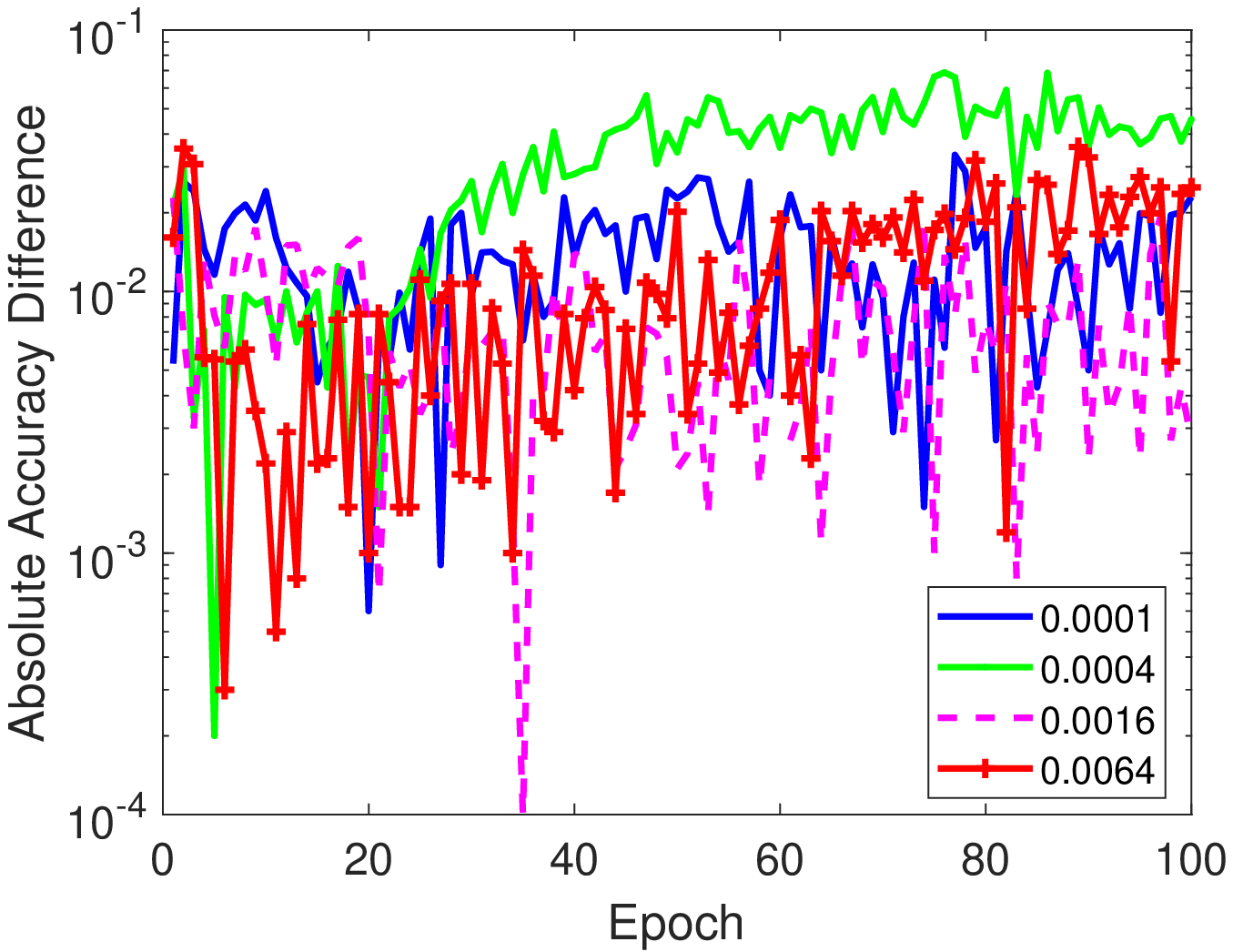}}
        \subfigure[Complete]{ \includegraphics[width=0.4\textwidth]{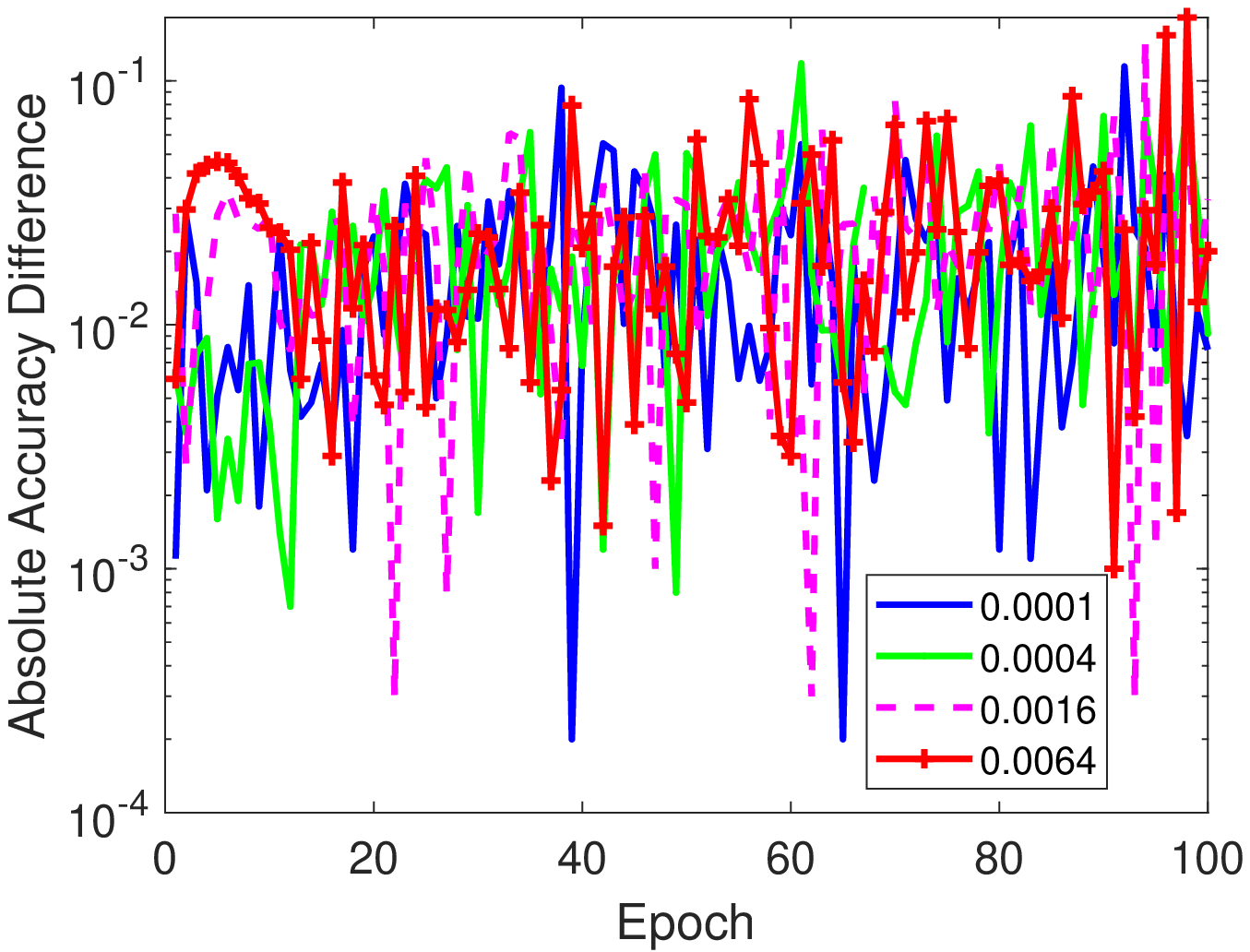}}
    \caption{\large Absolute accuracy difference versus epochs for nonconvex  model. The absolute accuracy difference performs similarly to the absolute loss difference.}
    \label{nonconvexacc}
\end{figure}

\section{Technical Lemmas}

\begin{lemma}\label{lemnum}
For any  $0<\lambda<1$ and $t\in \mathbb{Z}^+$, it holds
$$\sum_{j=0}^{t-1} \frac{\lambda^{t-1-j}}{j+1}\leq \frac{C_{\lambda}}{t}$$
with $C_{\lambda}:=\ln\frac{1}{\lambda}\frac{\lambda^{\ln\frac{1}{\lambda}}}{\lambda}+\frac{\ln^2\frac{1}{\lambda}}{16\lambda}\lambda^{\frac{\ln\frac{1}{\lambda}}{8}}+\frac{2}{\lambda \ln\frac{1}{\lambda}}$.
\end{lemma}

\begin{lemma}\label{recrusion}[Lemmas 2.5 and 3.7, \cite{hardt2015train}]
Fix an arbitrary sequence of updates $G_1,\dots,G_T$
and another sequence $G_1',\dots,G_T'.$
Let ${\bf x}^0={\bf y}^0$ be a starting point in $V$ and define
$\delta_t = \|{\bf x}^t- {\bf y}^t\|$ where ${\bf x}^t,  {\bf y}^t$ are defined recursively through
\begin{align*}
{\bf x}^{t+1}   = G_t({\bf x}^t),\,\,~~\,~~ {\bf y}^{t+1}  = G_t'( {\bf y}^t).
\end{align*}
Then, we have the recurrence relation
$\delta_0=0$,
\begin{align*}
\delta_{t+1} & \le \begin{cases}
\eta\delta_{t} & \text{$G_t=G_t'$ is $\eta$-expansive}  \\
\min(\eta,1) \delta_{t}+2\sigma_t & \text{$G_t$ and $G_t'$ are $\sigma$-bounded,}\,  \text{$G_t$ is $\eta$ expansive.}
\end{cases}
\end{align*}
For a nonnegative step size $\alpha\ge 0$ and a function $f\colon V\to\mathbb{R},$
we define  $G_{f,\alpha}$ as
$$G_{f,\alpha}({\bf x}) = \textbf{\emph{Proj}}_{V}( {\bf x} - \alpha \nabla f({\bf x})),$$
where $V$ is a closed convex set.
Assume that $f$ is $L$-smooth.
Then, the following properties hold.
\begin{itemize}
\item $G_{f,\alpha}$ is  $(1+\alpha L)$-expansive.

\item
If $f$ is convex. Then, for any $\alpha \le 2/L,$ the gradient update $G_{f,\alpha}$ is $1$-expansive.

\item If $f$ is $\nu$-strongly convex. Then, for $\alpha\leq\frac{1}{L}$,
$G_{f,\alpha}$ is  $\left(1- \alpha\nu \right)$-expansive.
\end{itemize}
\end{lemma}

\begin{lemma} \label{lemcore}
Let Assumption \ref{ass1} hold.
Assume two sample sets $S$ and $S'$ just differs at one sample in the first $n$ sample. And let ${\bf x}^T$ and ${\bf y}^T$ be the corresponding outputs of D-SGD applied to these two sets after $T$ steps.  Then, for every $\xi\sim \mathcal{D}$ and every
$t_0\in\{0,1,\dots,n\},$ under both the random update rule and the random
permutation rule, we have
$$
\EE\left|f({\bf x}^T;\xi)-f( {\bf y}^T;\xi)\right| \le \frac{t_0}{n}\sup_{{\bf x}\in V,\xi}f({\bf x};\xi)
+ B\EE\left[\delta_T\mid\delta_{t_0}=0\right].
$$
\end{lemma}

 \begin{lemma}\label{globalbound}
Given the stepsize   $(\alpha_t)_{t\geq 0}>0$ and assume $({\bf x}^{t}(i))_{t\geq 0}$ are generated by D-SGD for all $i\in\{1,2\ldots,m\}$. If  Assumption 3 holds, we have the following bound
\begin{align}
  \left[\sum_{i=1}^m\|{\bf x}^t(i)- {\bf x}^t\|^2\right]^{\frac{1}{2}}\leq  2\sqrt{m}B\sum_{j=0}^{t-1}\alpha_j\lambda^{t-1-j}.
\end{align}
\end{lemma}

\begin{lemma}\label{globalbound2}
Denote the matrix ${\bf X}^{t}:=\begin{bmatrix}
  {\bf x}^{t}(1),  {\bf x}^{t}(2),
    \ldots,
    {\bf x}^{t}(m)
\end{bmatrix}^{\top}\in\mathbb{R}^{m\times d}$. Assume the conditions of Lemma \ref{globalbound} hold, we then get
\begin{align}
 \|({\bf W}-{\bf P}){\bf X}^t\|\leq \sqrt{m}B\sum_{j=0}^{t-1}\alpha_j\lambda^{t-j}.
\end{align}
\end{lemma}
\section{Proofs of Technical Lemmas}
\subsection{Proof of Lemma \ref{lemnum}}
For any $x\in[j,j+1]$, $j+1\geq x$ and $\lambda^{t-1-j}\leq \lambda^{t-1-x}$, we then get
\begin{equation}\label{lemmumt1}
    \begin{aligned}
    &\sum_{j=0}^{t-1} \frac{\lambda^{t-1-j}}{j+1}=\lambda^{t-1}+    \sum_{j=1}^{t-1} \frac{\lambda^{t-1-j}}{j+1}\leq \lambda^{t-1}+\sum_{j=1}^{t-1}\int_{j}^{j+1}\frac{\lambda^{t-1-x}}{x}dx\\
    &=\lambda^{t-1}+ \lambda^{t-1}\int_{1}^{t}\frac{\lambda^{-x}}{x}dx.
    \end{aligned}
\end{equation}
We turn to the bound
 \begin{equation*}
    \begin{aligned}
     &\int_{1}^{t}\frac{\lambda^{-x}}{x}dx=  \int_{1}^{\frac{t}{2}}\frac{\lambda^{-x}}{x}dx+  \int_{\frac{t}{2}}^{t}\frac{\lambda^{-x}}{x}dx\\
     &\leq\lambda^{-\frac{t}{2}}\int_{1}^{\frac{t}{2}}\frac{1}{x}dx+\frac{2}{t}\int_{\frac{t}{2}}^{t}\lambda^{-x}dx\leq\lambda^{-\frac{t}{2}}\ln t+\frac{2\lambda^{- t}}{t\ln\frac{1}{\lambda}}
    \end{aligned}.
\end{equation*}
With \eqref{lemmumt1}, we are then led to
\begin{equation*}
    \begin{aligned}
    &\sum_{j=0}^{t-1} \frac{\lambda^{t-1-j}}{j+1}\leq  \lambda^{t-1}+ \lambda^{\frac{t}{2}-1}t+\frac{2}{t\lambda \ln\frac{1}{\lambda}},
    \end{aligned}
\end{equation*}
where we used the fact $\ln t\leq t$.
Now, we provide the bounds for  $\sup_{t\geq 1}\{t\lambda^{t-1}+\lambda^{\frac{t}{2}-1}t^2\}\leq \sup_{t\geq 1}\{t\lambda^{t-1}\}+\sup_{t\geq 1}\{\lambda^{\frac{t}{2}-1}t^2\}$.
Note that $\ln (t\lambda^{t-1})=\ln t+(t-1)\ln\lambda$, whose derivative is $\frac{1}{t}+\ln\lambda$. It is easy to check that $t=\ln\frac{1}{\lambda}$  achieves the maximum, which indicates
 $$\sup_{t\geq 1}\{t\lambda^{t-1}\}\leq\ln\frac{1}{\lambda}\frac{\lambda^{\ln\frac{1}{\lambda}}}{\lambda}.$$
 Similarly, we get
 $$\sup_{t\geq 1}\{\lambda^{\frac{t}{2}-1}t^2\}\leq \frac{\ln^2\frac{1}{\lambda}}{16\lambda}\lambda^{\frac{\ln\frac{1}{\lambda}}{8}}.$$
\subsection{Proof of Lemma \ref{lemcore}}
Proof of Lemma \ref{lemcore} is almost identical to the proof of Lemma 3.11 in \citep{hardt2015train}.
\subsection{Proof of Lemma \ref{globalbound}}
We denote that
$${\bf \zeta}^t:=\alpha_t\begin{bmatrix}
 \nabla f( {\bf x}^{t}(1);\xi_{j_t(1)}) ,  \nabla f( {\bf x}^{t}(2); \xi_{j_t(2)}),
    \ldots,
\nabla f( {\bf x}^{t}(m); \xi_{j_t(m)})
\end{bmatrix}^{\top}\in\mathbb{R}^{m\times d}.$$
With Assumption \ref{ass1}, $\|{\bf \zeta}^t\|\leq \alpha_t\sqrt{m}B$.
Then the global scheme can be presented as
\begin{align}\label{xtglobal}
{\bf X}^{t+1}=\PP_{V^m}\Big[{\bf W}{\bf X}^{t}-{\bf \zeta}^t\Big],
\end{align}
where $V^m:=\underbrace{V\oplus V\ldots \oplus V}_{m}$.
Noticing the fact
\begin{equation}\label{trans}
    {\bf W} {\bf P}= {\bf P} {\bf W}={\bf P}.
\end{equation}
With Lemma \ref{mi}, we have
$$\|{\bf W}-{\bf P}\|_{\textrm{op}}\leq \lambda.$$
Multiplication of  both sides of \eqref{xtglobal} with  ${\bf P}$ together with \eqref{trans} then tells us
\begin{equation}\label{xtglobal3}
\begin{aligned}
&(\mathbb{I}-{\bf P}){\bf X}^{t+1}=(\mathbb{I}-{\bf P})\PP_{V^m}\Big[{\bf W}{\bf X}^{t}-{\bf \zeta}^t\Big]\\
&=\PP_{V^m}\Big[{\bf W}{\bf X}^{t}-{\bf \zeta}^t\Big]-{\bf P}\cdot\PP_{V^m}\Big[{\bf W}{\bf X}^{t}\Big]+{\bf P}\cdot\PP_{V^m}\Big[{\bf W}{\bf X}^{t}\Big]-{\bf P}\cdot\PP_{V^m}\Big[{\bf W}{\bf X}^{t}-{\bf \zeta}^t\Big].
\end{aligned}
\end{equation}
It is easy to check that ${\bf W}{\bf X}^{t}\in V^m$ and ${\bf P}{\bf W}{\bf X}^{t}\in V^m$. Thus, it follows ${\bf P}\cdot\PP_{V^m}\Big[{\bf W}{\bf X}^{t}\Big]={\bf P} {\bf W}{\bf X}^{t}=\PP_{V^m}\Big[{\bf P}{\bf W}{\bf X}^{t}\Big]$.
From \eqref{xtglobal3}, letting $\mathbb{I}\in \RR^m$ be the identical matrix,
\begin{equation}\label{xtglobalineq}
\begin{aligned}
&\|(\mathbb{I}-{\bf P}){\bf X}^{t+1}\|\leq \left\|\PP_{V^m}\Big[{\bf W}{\bf X}^{t}-{\bf \zeta}^t\Big]- \PP_{V^m}\Big[{\bf P}{\bf W}{\bf X}^{t}\Big]\right\|+\|{\bf \zeta}^t\|\\
&\qquad\leq \|{\bf W}{\bf X}^{t}-{\bf \zeta}^t-{\bf P}{\bf W}{\bf X}^{t}\|+\sqrt{m}B\alpha_t\leq \|{\bf W}{\bf X}^{t} -{\bf P}{\bf W}{\bf X}^{t}\|+2\sqrt{m}B\alpha_t\\
&\qquad\overset{a)}{\leq} \|({\bf W}-{\bf P})(\mathbb{I}-{\bf P}){\bf X}^{t}\|+2\sqrt{m}B\alpha_t\overset{b)}{\leq}\lambda\|(\mathbb{I}-{\bf P}){\bf X}^{t}\|+2\sqrt{m}B\alpha_t.
\end{aligned}
\end{equation}
where $a)$ uses the fact $({\bf W}-{\bf P})(\mathbb{I}-{\bf P})={\bf W}-2{\bf P}+{\bf W}{\bf P}={\bf W}- {\bf P}={\bf W}- {\bf P}W$, and $b)$ depends on Lemma \ref{mi}.
Then, we derive
\begin{align}\label{xtglobal4}
\|{\bf X}^{t}-{\bf P}{\bf X}^{t}\|\leq 2\sqrt{m}B\sum_{j=0}^{t-1}\alpha_j\lambda^{t-1-j}.
\end{align}
where we used initialization ${\bf X}^{0}=\textbf{0}$.
\subsection{Proof of Lemma \ref{globalbound2}}
Direct computations give
\begin{equation*}
  \begin{aligned}
  ({\bf W}-{\bf P}){\bf X}^{t+1}&=({\bf W}-{\bf P}){\bf W}{\bf X}^t-({\bf W}-{\bf P}){\bf \zeta}^t\\
  &=({\bf W}-{\bf P})({\bf W}-{\bf P}){\bf X}^t-({\bf W}-{\bf P}){\bf \zeta}^t,
  \end{aligned}
\end{equation*}
 where we used $(W-{\bf P})W=(W-{\bf P})(W-{\bf P})$. Thus, we obtain
 \begin{equation*}
  \begin{aligned}
  \|(W-{\bf P}){\bf X}^{t+1}\|\leq \lambda\|(W-{\bf P}){\bf X}^t\|+\sqrt{m}\lambda B\alpha_t,
  \end{aligned}
\end{equation*}

\section{Proof of Theorem \ref{th1}}
Without loss of generalization , assume two sample sets $S$ and $S'$ just differs at one sample in the first $m$ sample. And let ${\bf x}^t$ and ${\bf y}^t$ be the corresponding outputs of D-SGD applied to these two sets.   Denoting that
$$\delta_{t}:=\|{\bf x}^t-{\bf y}^t\|.$$
At iteration $t$, with probability $1-\frac{1}{n}$,
 \begin{equation}\label{dec1}
     \begin{aligned}
    {\bf x}^{t+1}- {\bf y}^{t+1}&=\frac{\sum_{i=1}^m\left[ \textbf{Proj}_{V}[ \sum_{l\in \mathcal{N}(i)} w_{i,l}{\bf x}^t(l)- \alpha_t\nabla f({\bf x}^t(i);\xi_{j_t(i)})]-\textbf{Proj}_{V}[\sum_{l\in \mathcal{N}(i)} w_{i,l} {\bf y}^{t}(l)-\alpha_t\nabla f( {\bf y}^{t}(i);\xi_{j_t(i)})] \right]}{m}\\
    &=\underbrace{\frac{\sum_{i=1}^m\left[ \textbf{Proj}_{V}[{\bf x}^t- \alpha_t\nabla f({\bf x}^t;\xi_{j_t(i)})]-\textbf{Proj}_{V}[ {\bf y}^{t}-\alpha_t\nabla f({\bf y}^{t};\xi_{j_t(i)})] \right]}{m}}_{\S}+\wp_t.
 \end{aligned}
 \end{equation}
 With gradient smooth assumption,
  \begin{equation}\label{bound1}
     \begin{aligned}
 \|\wp_t\|&\leq \frac{\sum_{i=1}^m ( \sum_{l\in \mathcal{N}(i)} w_{i,l}\|{\bf x}^t(l)-{\bf x}^t\| +\alpha_t B\|{\bf x}^t(i)-{\bf x}^t\|)}{m}+ \frac{\sum_{i=1}^m (  \sum_{l\in \mathcal{N}(i)} w_{i,l}\|{\bf y}^t(l)-{\bf y}^t\| +\alpha_t B\|{\bf y}^t(i)-{\bf y}^t\|)}{m}\\
 &\leq \frac{\sum_{i=1}^m ( \sum_{l=1}^m w_{i,l}\|{\bf x}^t(l)-{\bf x}^t\| +\alpha_t B\|{\bf x}^t(i)-{\bf x}^t\|)}{m}+ \frac{\sum_{i=1}^m (  \sum_{l=1}^m w_{i,l}\|{\bf y}^t(l)-{\bf y}^t\| +\alpha_t B\|{\bf y}^t(i)-{\bf y}^t\|)}{m}\\
  &\leq \frac{\sum_{l=1}^m \|{\bf x}^t(l)-{\bf x}^t\| +\alpha_t  B\sum_{i=1}^m  \|{\bf x}^t(i)-{\bf x}^t\|}{m}+ \frac{  \sum_{l=1}^m \|{\bf y}^t(l)-{\bf y}^t\| +\alpha_t B\sum_{i=1}^m\|{\bf y}^t(i)-{\bf y}^t\|}{m}\\
 &\leq \frac{ ( 1+\alpha_t B)\sqrt{m}[\sum_{i=1}^m\|{\bf x}^t(i)-{\bf x}^t\|^2]^{\frac{1}{2}}}{m}+ \frac{ ( 1+\alpha_t B)\sqrt{m}[\sum_{i=1}^m\|{\bf y}^t(i)-{\bf y}^t\|^2]^{\frac{1}{2}}}{m}\\
 &\leq 4( 1+\alpha_t B)B\sum_{j=0}^{t-1}\alpha_j\lambda^{t-1-j}.
 \end{aligned}
 \end{equation}
 Due to the convexity, $\textbf{Proj}_{V}[\cdot- \alpha_t\nabla f(\cdot;\xi_{j_t(i)})]$ is 1-expansive when $\alpha_t\leq\frac{2}{L}$. Thus, we have
 $$\|\S\|\leq \|{\bf x}^t-{\bf y}^t\|=\delta_t.$$
 With probability $\frac{1}{n}$,
 \begin{equation}\label{dec2}
     \begin{aligned}
    {\bf x}^{t+1}-{\bf y}^{t+1}&=\frac{\sum_{i=2}^m\left[ \textbf{Proj}_{V}[\sum_{l\in \mathcal{N}(i)} w_{i,l}{\bf x}^t(l)- \alpha_t\nabla f({\bf x}^t;\xi_{j_t(i)})]-\textbf{Proj}_{V}[ \sum_{l\in \mathcal{N}(i)} w_{i,l}{\bf y}^{t}(l)-\alpha_t\nabla f({\bf y}^{t};\xi_{j_t(i)})] \right]}{m}\\
    &+\frac{ \left[ \textbf{Proj}_{V}[\sum_{l\in \mathcal{N}(1)} w_{1,l}{\bf x}^t(l)- \alpha_t\nabla f({\bf x}^t;\xi'_{j_t(1)})]-\textbf{Proj}_{V}[\sum_{l\in \mathcal{N}(1)} w_{1,l} {\bf y}^{t}(l)-\alpha_t\nabla f( {\bf y}^{t};\xi_{j_t(1)})] \right]}{m}+\Im_t\\
    &=\underbrace{\frac{\sum_{i=2}^m\left[ \textbf{Proj}_{V}[{\bf x}^t- \alpha_t\nabla f({\bf x}^t;\xi_{j_t(i)})]-\textbf{Proj}_{V}[ {\bf y}^{t}-\alpha_t\nabla f({\bf y}^{t};\xi_{j_t(i)})] \right]}{m}}_{\sharp}\\
    &+\frac{ \left[ \textbf{Proj}_{V}[{\bf x}^t-\alpha_t\nabla f({\bf x}^t;\xi'_{j_t(1)})]-\textbf{Proj}_{V}[ {\bf y}^{t}-\alpha_t\nabla f( {\bf y}^{t};\xi_{j_t(1)})] \right]}{m}+\Im_t.
 \end{aligned}
 \end{equation}
 It is easy to check that $\|\Im_t\|\leq 4( 1+\alpha_t B)B\sum_{j=0}^{t-1}\alpha_j\lambda^{t-1-j}$. With  Lemma \ref{recrusion}, we have
 $$\|\sharp\|\leq \frac{m-1}{m}\delta_t,~\,~\,~\left\|\frac{ \left[ \textbf{Proj}_{V}[{\bf x}^t- \alpha_t\nabla f({\bf x}^t;\xi_{j'_t(1)})]-\textbf{Proj}_{V}[{\bf y}^{t}-\alpha_t\nabla f({\bf y}^{t};\xi_{j_t(1)})] \right]}{m}\right\|\leq\frac{\delta_t+2B\alpha_t}{m}.$$
 Thus, we derive
 $$\EE\delta_{t+1}\leq \EE\delta_{t}+\frac{2B\alpha_t}{mn}+4( 1+\alpha_t B)B\sum_{j=0}^{t-1}\alpha_j\lambda^{t-1-j}.$$
 That also means
 $$\EE\delta_{T}\leq   \frac{2B\sum_{t=0}^{T-1}\alpha_t}{mn}+4B\sum_{t=0}^{T-1}( 1+\alpha_t B)\sum_{j=0}^{t-1}\alpha_j\lambda^{t-1-j}.$$
 Noticing the fact
 \begin{equation}\label{lips}
\EE|f({\bf x}^T;\xi)-f( {\bf y}^T;\xi)|\leq B\EE\|{\bf x}^T-{\bf y}^T\|=B\EE\delta_T,
 \end{equation}
 we then proved the result.
  \section{Proof of Proposition  \ref{pro1}}
  Let ${\bf x}^t$ and ${\bf y}^t$ be the output of D-SGD with only one different sample after $t$ iterations.
   We have
   \begin{align*}
 \| \textrm{ave}({\bf x}^T)-\textrm{ave}({\bf y}^T)\|=\left\|\frac{\sum_{t=1}^{T-1}\alpha_t({\bf x}^t-{\bf y}^t)}{\sum_{t=1}^{T-1}\alpha_t}\right\|\leq  \frac{\sum_{t=1}^{T-1}\alpha_t\|{\bf x}^t-{\bf y}^t\|}{\sum_{t=1}^{T-1}\alpha_t}.
   \end{align*}
   As the same proofs in Theorem \ref{th1}, we can derive
   $$\|{\bf x}^t-{\bf y}^t\|\leq \frac{2B\sum_{k=1}^{t-1}\alpha_k}{mn}+4B\sum_{k=1}^{t-1}( 1+\alpha_k B)\sum_{j=0}^{k-1}\alpha_j\lambda^{k-1-j}.$$

   1) when $\alpha_t\equiv\alpha$, we have
    $$\|{\bf x}^t-{\bf y}^t\|\leq \frac{2B\alpha(t-1)}{mn}+\frac{4\alpha B( 1+\alpha B)(t-1)}{1-\lambda}1_{\lambda\neq 1}.$$
    Thus, we get
      \begin{align*}
 \| \textrm{ave}({\bf x}^T)-\textrm{ave}({\bf y}^T)\|\leq \frac{B\alpha(t-1)}{mn}+\frac{2\alpha B( 1+\alpha B)(t-1)}{1-\lambda}1_{\lambda\neq 1}.
   \end{align*}

      2) when $\alpha_t=\frac{1}{t+1}$, we have
    $$\|{\bf x}^t-{\bf y}^t\|\leq \frac{2B\ln t}{mn}+\frac{4\alpha B( 1+B)C_{\lambda}}{t}1_{\lambda\neq 1}.$$
    Thus, we get
      \begin{align*}
 \| \textrm{ave}({\bf x}^T)-\textrm{ave}({\bf y}^T)\|\leq \frac{B\ln T}{mn}+\frac{4\alpha B( 1+B)}{\ln (T+1)}1_{\lambda\neq 1}.
   \end{align*}

   The Lipschitz continuity of the loss function then proves the result.
 \section{Proof of Theorem \ref{th2}}
 Without loss of generalization , we assume $\lambda\neq 0$.
 Due to the strong convexity with $\nu$, $\textbf{Proj}_{V}[\cdot- \alpha_t\nabla f(\cdot;\xi_{j_t(i)})]$ is $(1- \alpha_t\nu)$-contractive when $\alpha_t\leq\frac{1}{L}$. Thus, in \eqref{dec1}, it holds
  $$\|\S\|\leq (1- \alpha_t\nu)\|{\bf x}^t-\overline{{\bf x}}^t\|=(1- \alpha_t\nu)\delta_t;$$
  and in \eqref{dec2},
   $$\|\sharp\|\leq \frac{m-1}{m}(1- \alpha_t\nu)\delta_t,$$~
   $$\left\|\frac{ \left[ \textbf{Proj}_{V}[{\bf x}^t- \alpha_t\nabla f({\bf x}^t;\xi_{j'_t(1)})]-\textbf{Proj}_{V}[ {\bf y}^{t}-\alpha_t\nabla f({\bf y}^{t};\xi_{j_t(1)})] \right]}{m}\right\|\leq\frac{(1- \alpha_t\nu)\delta_t+2B\alpha_t}{m}.$$
   Therefore, we can get
          \begin{equation}   \label{finals}
       \begin{aligned}
\EE\delta_{t+1}\leq (1- \alpha_t\nu)\EE\delta_{t}+\frac{2B\alpha_t}{mn}+4( 1+\alpha_t B)B\sum_{j=0}^{t-1}\lambda^{t-1-j}\alpha_j.
       \end{aligned}
    \end{equation}

\textbf{a)} When $0<\lambda<1$, $\sum_{j=0}^{t-1}\lambda^{t-1-j}\leq \frac{1}{1-\lambda}$ and $\alpha_t\equiv\alpha$. Then we are led to
       \begin{equation*}
       \begin{aligned}
\EE\delta_{t+1}\leq (1- \alpha\nu)\EE\delta_{t}+\frac{2B\alpha}{mn}+4( 1+\alpha B)B\alpha \frac{1}{1-\lambda}.
       \end{aligned}
    \end{equation*}
    The recursion of inequality gives us
    \begin{equation*}
       \begin{aligned}
       \EE\delta_{T}\leq \sum_{t=0}^{T-1}(1- \alpha\nu)^t\left[\frac{2B\alpha}{mn}+4( 1+\alpha B)B\alpha \frac{1}{1-\lambda} \right]=\frac{2B}{mn\nu}+\frac{4( 1+\alpha B)B}{\nu} \frac{1}{1-\lambda}.
       \end{aligned}
    \end{equation*}
      Once using \eqref{lips}, we then get the result.\\

\textbf{b)} In \eqref{finals}, by replacing $\alpha$ with $\alpha_t=\frac{1}{\nu (t+1)}$,
       \begin{equation}
       \begin{aligned}
\EE\delta_{t+1}\leq (1- \frac{1}{t+1})\EE\delta_{t}+\Big[\frac{2B}{\nu mn}+4( 1+ \frac{ B}{\nu})\frac{ B}{\nu}  \frac{1_{\lambda\neq 0}}{1-\lambda}\Big]\frac{1}{t+1}.
       \end{aligned}
    \end{equation}
    Thus, we get
       \begin{equation}
       \begin{aligned}
\EE\delta_{t}\leq \frac{2B}{\nu mn}+ 4( 1+ \frac{ B}{\nu})\frac{ B}{\nu}  \frac{1_{\lambda\neq 0}}{1-\lambda}.
       \end{aligned}
    \end{equation}

    \section{Proof of Theorem \ref{th3}}
Assume $t_0\in \{1,2,\ldots,n\}$ to be determined, with Lemma \ref{lemcore},
$$
\EE\left|f({\bf x}^T;\xi)-f( {\bf y}^T;\xi)\right| \le \frac{t_0}{n}
+ L\EE\left[\delta_T\mid\delta_{t_0}=0\right],
$$
where we used $\sup_{{\bf x}\in V,\xi}f({\bf x};\xi)\leq 1$. In the following, we consider the case $\delta_{t_0}=0$.
With Lemma \ref{recrusion}, in \eqref{dec1},
  $$\|\S\|\leq (1+ \alpha_tL)\|{\bf x}^t-{\bf y}^t\|=(1+ \alpha_tL)\delta_t;$$
  and in \eqref{dec2},
   $$\|\sharp\|\leq \frac{m-1}{m}(1+ \alpha_tL)\delta_t,$$$$~\left\|\frac{ \left[ \textbf{Proj}_{V}[{\bf x}^t- \alpha_t\nabla f({\bf x}^t;\xi_{j'_t(1)})]-\textbf{Proj}_{V}[{\bf y}^{t}-\alpha_t\nabla f({\bf y}^{t};\xi_{j_t(1)})] \right]}{m}\right\|\leq\frac{\delta_t+2B\alpha_t}{m}.$$
   Therefore,  when $t>t_0$, we can get
\begin{equation*}
    \begin{aligned}
    \EE(\delta_{t+1}\mid\delta_{t_0}=0)&\leq (1-\frac{1}{n})(1+ \alpha_tL)\EE(\delta_{t}\mid\delta_{t_0}=0)+\frac{1}{n}\Big[\frac{m-1}{m}(1+ \alpha_tL)+\frac{1}{m}\Big]\delta_t\\
    &+\frac{2B\alpha_t}{mn}+4( 1+\alpha_t B)B\sum_{j=0}^{t-1}\alpha_j\lambda^{t-1-j}\\
    &=\Big[1+(1-\frac{1}{mn})\alpha_tL\Big]\EE(\delta_{t}\mid\delta_{t_0}=0)+\frac{2B\alpha_t}{mn}+4( 1+\alpha_t B)B\sum_{j=0}^{t-1}\alpha_j\lambda^{t-1-j}.
    \end{aligned}
\end{equation*}
With $\alpha\leq \frac{c}{t}$ and Lemma \ref{lemnum}, we are then led to
\begin{equation*}
    \begin{aligned}
    \EE(\delta_{t+1}\mid\delta_{t_0}=0)&\leq \Big[1+cL(1-\frac{1}{mn})/t\Big]\EE(\delta_{t}\mid\delta_{t_0}=0)+(\frac{2Bc}{mn}+4( 1+c B)BC_{\lambda})/t\\
    &\leq\exp(cL(1-\frac{1}{mn})/t)\EE(\delta_{t}\mid\delta_{t_0}=0)+(\frac{2Bc}{mn}+4( 1+c B)BC_{\lambda})/t.
    \end{aligned}
\end{equation*}
That then yields
\begin{equation*}
    \begin{aligned}
    \EE(\delta_{T}\mid\delta_{t_0}=0)&\leq \sum_{t=t_0+1}^T \Big[\exp(\sum_{i=t+1}^T cL(1-\frac{1}{mn})/i)\Big] (\frac{2Bc}{mn}+4( 1+c B)BC_{\lambda})/t\\
    &\leq  \sum_{t=t_0+1}^T \Big[\exp( cL(1-\frac{1}{mn})\ln\frac{T}{t})\Big] (\frac{2Bc}{mn}+4( 1+c B)BC_{\lambda})/t\\
    &=(\frac{2Bc}{mn}+4( 1+c B)BC_{\lambda})T^{cL(1-\frac{1}{mn})}\sum_{t=t_0+1}^T \frac{1}{t^{cL(1-\frac{1}{mn})+1}}\\
    &\leq(\frac{2Bc}{mn}+4( 1+c B)BC_{\lambda})T^{cL(1-\frac{1}{mn})}  \frac{cL(1-\frac{1}{mn})}{t_0^{cL(1-\frac{1}{mn})}}\\
    &\leq (\frac{2Bc}{mn}+4( 1+c B)BC_{\lambda}) cL  \frac{T^{cL} }{t_0^{cL}}
    \end{aligned}
\end{equation*}
From Lemma \ref{lemcore}, we get
$$
\EE\left|f({\bf x}^T;\xi)-f({\bf y}^T;\xi)\right| \leq \frac{t_0}{mn}
+ B cL(\frac{2Bc}{mn}+4( 1+c B)BC_{\lambda}) \frac{T^{cL} }{t_0^{cL}}.
$$
Setting $t_0=c^{\frac{1}{1+cL}}T^{\frac{cL}{1+cL}}$, when $c$ is small enough, $t_0\leq n$. We then have
$$\EE\left|f({\bf x}^T;\xi)-f({\bf y}^T;\xi)\right| \leq \frac{c^{\frac{1}{1+cL}}T^{\frac{cL}{1+cL}}}{mn}
+ B Lc^{\frac{1}{1+cL}}(\frac{2Bc}{mn}+4( 1+c B)BC_{\lambda})T^{\frac{cL}{1+cL}}.$$

\section{Proof of Lemma \ref{th-1}}
Without loss of generalization , we assume $\lambda\neq 0$.
 With the projection ${\bf x}^*=\textbf{Proj}_{V}({\bf x}^*)$, it then follows
\begin{equation}\label{th0-t1}
\begin{aligned}
    \EE\|{\bf x}^{t+1}-{\bf x}^*\|^2&\leq  \EE\left\|\sum_{i=1}^m \textbf{Proj}_{V}[\sum_{l\in \mathcal{N}(i)} w_{i,l} {\bf x}^{t}(l)-\alpha_t\nabla f( {\bf x}^{t}(i); \xi_{j_t(i)} )]/m-\textbf{Proj}_{V}({\bf x}^*)\right\|^2\\
    &=\EE\left\|\Big[\sum_{i=1}^m \textbf{Proj}_{V}[\sum_{l\in \mathcal{N}(i)} w_{i,l} {\bf x}^{t}(l)-\alpha_t\nabla f( {\bf x}^{t}(i); \xi_{j_t(i)} )]-\textbf{Proj}_{V}({\bf x}^*)\Big]/m\right\|^2\\
    &\leq \frac{1}{m}\EE\|{\bf W}{\bf X}^{t}-{\bf \zeta}^t-{\bf X}^*\|^2=\frac{1}{m}\EE\|{\bf P}{\bf X}^{t}-{\bf X}^*-\alpha_t{\bf \zeta}^t+({\bf W}-{\bf P}){\bf X}^{t}\|^2\\
    &=\EE\|{\bf x}^{t}-{\bf x}^*\|^2+\underbrace{\frac{2}{m}\EE\langle({\bf W}-{\bf P}){\bf X}^{t}-{\bf \zeta}^t,{\bf P}{\bf X}^{t}-{\bf X}^*\rangle}_{\textrm{I}}+\underbrace{\frac{1}{m}\EE\|\alpha_t{\bf \zeta}^t-({\bf W}-{\bf P}){\bf X}^{t}\|^2}_{\textrm{II}},
\end{aligned}
\end{equation}
where ${\bf X}^*:=\begin{bmatrix}
 {\bf x}^*,  {\bf x}^*,
    \ldots,
 {\bf x}^*
\end{bmatrix}^{\top}\in\mathbb{R}^{m\times d}$. Now, we bound $\textrm{I}$ and $\textrm{II}$:  we first observe
\begin{equation*}
    \begin{aligned}
    \langle({\bf W}-{\bf P}){\bf X}^{t},{\bf P}{\bf X}^{t}-{\bf X}^*\rangle=\langle{\bf P}({\bf W}-{\bf P}){\bf X}^{t},{\bf X}^{t}\rangle-\langle{\bf X}^{t},({\bf W}-{\bf P}){\bf X}^*\rangle=0.
        \end{aligned}
\end{equation*}
The convexity and Lemma \ref{globalbound} tell us
\begin{equation*}
\begin{aligned}
    &\frac{2}{m}\EE\langle({\bf W}-{\bf P}){\bf X}^{t}-{\bf \zeta}^t,{\bf P}{\bf X}^{t}-{\bf X}^*\rangle\leq   \frac{2}{m}\EE\langle-{\bf \zeta}^t,{\bf P}{\bf X}^{t}-{\bf X}^*\rangle  +\frac{2}{m}\EE\langle({\bf W}-{\bf P}){\bf X}^{t},{\bf P}{\bf X}^{t}-{\bf X}^*\rangle\\
    &\leq \sum_{i=1}^m \frac{2\alpha_t}{m}\EE\langle -\nabla f({\bf x}(i)),  {\bf x}^{t}-{\bf x}^*\rangle\\
    &\leq \sum_{i=1}^m \frac{2\alpha_t}{m}\EE\langle -\nabla f({\bf x}^{t}), {\bf x}^{t}-{\bf x}^*\rangle+\sum_{i=1}^m \frac{2\alpha_t}{m}\EE\langle \nabla f({\bf x}^{t})-\nabla f({\bf x}(i)),
      {\bf x}^{t}-{\bf x}^*\rangle\\
      &\leq -2\alpha_t \EE(f({\bf x}^t)-f({\bf x}^*))+\frac{4\alpha_t Lr}{\sqrt{m}}\EE(\sum_{i=1}^m\|{\bf x}^{t}-{\bf x}(i)\|^2)^{\frac{1}{2}}\\
      &\leq -2\alpha_t\EE(f({\bf x}^t)-f({\bf x}^*))+8\alpha_t LrB\sum_{j=0}^{t-1}\alpha_j\lambda^{t-1-j}.
    \end{aligned}
 \end{equation*}
  And with Lemma \ref{globalbound2}, we  have
 \begin{equation}\label{th-1-t1}
\begin{aligned}
    \frac{1}{m}\EE\|\alpha_t{\bf \zeta}^t-({\bf W}-{\bf P}){\bf X}^{t}\|^2 \leq2\alpha_t^2B^2 +\frac{2\lambda^2(B\sum_{j=0}^{t-1}\alpha_j\lambda^{t-1-j})^2}{m}.
    \end{aligned}
 \end{equation}
 Thus, we then get
  \begin{equation*}
\begin{aligned}
 &\frac{\sum_{t=0}^{T-1} \alpha_t\EE(f({\bf x}^t)-f({\bf x}^*))}{\sum_{t=0}^{T-1}}\leq \frac{\|{\bf x}^0-{\bf x}^*\|^2}{\sum_{t=0}^{T-1}\alpha_t}+\frac{8 LrB\sum_{t=0}^{T-1}\alpha_t\sum_{j=0}^{t-1}\alpha_j\lambda^{t-1-j}}{\sum_{t=0}^{T-1}\alpha_t}\\
 &+\frac{2\lambda^2B^2\sum_{t=0}^{T-1}\alpha_t(\sum_{j=0}^{t-1}\alpha_j\lambda^{t-1-j})^2}{m\sum_{t=0}^{T-1}\alpha_t}+\frac{2B^2\sum_{t=0}^{T-1}\alpha_t^2}{m\sum_{t=0}^{T-1}\alpha_t}.
    \end{aligned}
 \end{equation*}
 The convexity then completes the proof.
\section{Proof of Lemma \ref{th0}}
We start from \eqref{th0-t1}
and bound $\textrm{I}$ and $\textrm{II}$: the strong convexity and Lemma \ref{globalbound} yields
\begin{equation*}
\begin{aligned}
    &\frac{2}{m}\EE\langle({\bf W}-{\bf P}){\bf X}^{t}-{\bf \zeta}^t,{\bf P}{\bf X}^{t}-{\bf X}^*\rangle\leq \sum_{i=1}^m \frac{2\alpha_t}{m}\EE\langle -\nabla f({\bf x}(i)),  {\bf x}^{t}-{\bf x}^*\rangle\\
    &\leq \sum_{i=1}^m \frac{2\alpha_t}{m}\EE\langle -\nabla f({\bf x}^{t}), {\bf x}^{t}-{\bf x}^*\rangle+\sum_{i=1}^m \frac{2\alpha_t}{m}\EE\langle \nabla f({\bf x}^{t})-\nabla f({\bf x}(i)),
      {\bf x}^{t}-{\bf x}^*\rangle\\
      &\leq -2\alpha_t\nu\EE\|{\bf x}^{t} - {\bf x}^*\|^2+\frac{4\alpha_t Lr}{\sqrt{m}}\EE(\sum_{i=1}^m\|{\bf x}^{t}-{\bf x}(i)\|^2)^{\frac{1}{2}}\\
      &\leq -2\alpha_t\nu\EE\|{\bf x}^{t} - {\bf x}^*\|^2+8\alpha_t LrB\sum_{j=0}^{t-1}\alpha_j\lambda^{t-1-j}.
    \end{aligned}
 \end{equation*}
Note that \eqref{th-1-t1} still holds in the strong convexity case.

a) Applying the bounds of $\textrm{I}$ and $\textrm{II}$ to \eqref{th0-t1}, letting $\alpha_t\equiv\alpha$,
\begin{equation*}
\begin{aligned}
    \EE\|{\bf x}^{t+1}-{\bf x}^*\|^2\leq  (1-2\alpha\nu)\EE\|{\bf x}^{t}-{\bf x}^*\|^2+\frac{8\alpha^2LrB}{1-\lambda}+\frac{2\lambda^2B^2\alpha^2}{m(1-\lambda)^2}.
\end{aligned}
\end{equation*}
Thus, we derive
\begin{equation*}
\begin{aligned}
    \EE\|{\bf x}^{T}-{\bf x}^*\|^2\leq   (1-2\alpha\nu)^T\EE\|{\bf x}^{0}-{\bf x}^*\|^2 +\frac{\frac{8\alpha^2LrB}{1-\lambda}+\frac{2\lambda^2B^2\alpha^2}{m(1-\lambda)^2}}{2\alpha\nu}.
\end{aligned}
\end{equation*}

b) Applying the bounds of $\textrm{I}$ and $\textrm{II}$ to \eqref{th0-t1}, letting $\alpha_t=\frac{1}{2\nu(t+1)}$, we get
\begin{equation*}
\begin{aligned}
    \EE\|{\bf x}^{t+1}-{\bf x}^*\|^2\leq  (1-\frac{1}{t+1})\EE\|{\bf x}^{t}-{\bf x}^*\|^2+\frac{4D_{\lambda}}{(t+1)^2}.
\end{aligned}
\end{equation*}
Thus, we derive
\begin{equation*}
\begin{aligned}
    \EE\|{\bf x}^{T}-{\bf x}^*\|^2\leq   \frac{\EE\|{\bf x}^{0}-{\bf x}^*\|^2}{T}+\frac{D_{\lambda}}{T}\sum_{t=2}^T\frac{1}{t}\leq \frac{\EE\|{\bf x}^{0}-{\bf x}^*\|^2}{T}+\frac{D_{\lambda}\ln T}{T}.
\end{aligned}
\end{equation*}
\section{Proofs of Theorem \ref{th4} and \ref{th5}}
Using the fact
$\epsilon_{\textrm{ex-gen}}\leq \epsilon_{\textrm{gen}}+\EE_{S,\mathcal{A}}[R_{S}(\mathcal{A}(S))-R_{S}(\overline{{\bf x}})]$
and noticing that $\EE_{S,\mathcal{A}}[R_{S}(\mathcal{A}(S))-R_{S}(\overline{{\bf x}})$ is computational error, we then get the result by proved results given above.
\end{document}